# FON: A Draft–Revise Dual-Organ Language Model without Sequence-Level Quadratic Attention

Helong Hu[1,*], Hongdan Pan[2], Shuiqing Hu[3]

[1] *Guangdong Ocean University, Zhanjiang, Guangdong 524088, China.* [2] *Maoming Administrative Service Center, Maoming, Guangdong 525200, China.* [3] *China Pacific Insurance (Group) Co., Ltd. Foshan Branch, Foshan, Guangdong, China.*

> **Corresponding author:** Helong Hu (also first author), Guangdong Ocean University, No.1 Haida Road, Mazhang District, Zhanjiang, Guangdong 524088, China. Email: 604232736@qq.com

Self-attention assigns two functions to one mechanism—where to read and how to mix—at the price of $O(L^2)$ training, an O(T) decode state, and position signals bound to the training length. Existing attention-free alternatives all optimize the single function of generation. We propose FON (Fusion-Organ Network), which differentiates one weight stack into a drafting organ (multi-tap learnable delays + content-adaptive recursion; causal, O(S) training, O(1) decode state, no sequence-level quadratic attention) and a revision organ that re-processes the written prefix through multi-scale ripple kernels injected via weak gates, never generating. At 30M parameters, iso-parameter and iso-token budget, FON reaches perplexity 131.03 on WikiText-103, beating Transformer (197.73) and Mamba2 (152.53); the advantage persists under all depth-aligned baselines and at the 70M tier (−20.8%). At 8× length extrapolation FON degrades by ~8–9% without positional-encoding engineering, between ALiBi and RoPE. The revision channel provides a structural pathway for in-domain infilling that causal architectures lack in principle (+19% over on-demand coupling), at a perplexity cost indistinguishable within measurement precision. A fill-mode control on the synthetic battery shows it is a dual-mode (generate + fill) test that the causal Transformer fails outright—FON-PC exceeds it on all six tasks—so FON is the only architecture here unifying causal generation with bidirectional fill. We report all costs and boundaries: impaired long-range retrieval relative to a non-generative bidirectional encoder, weaker out-of-domain code infilling than a same-recipe Transformer, and a 6.5–12.3× wall-clock gap.

---

## 1 Introduction

### 1.1 Structural constraints of the attention paradigm: from engineering costs to functional boundaries

Self-attention [1,2] uses a single mechanism both to choose where to read information and to decide how to mix it, implemented as all-pairs association over a two-dimensional matrix X ∈ R^(L×d). This design is an engineering triumph, but its structural costs become increasingly visible with scale and workload: (1) training complexity $O(L^2)$, with attention dominating compute at long sequence lengths; (2) decode state O(T)—the KV cache grows linearly with generation length [6,7], making the memory wall of long-context deployment a first-order constraint; (3) positional signals are bound to the training length, so performance collapses beyond it and can only be partially repaired by positional-encoding engineering (RoPE [31], ALiBi [32], positional interpolation [33], YaRN [34], NoPE [35]).

This paper unifies these constraints under a computational-geometry view: attention flattens sequence modeling into a two-dimensional computational plane—each position is processed exactly once along the time axis; to process is to finalize. All three constraints follow from this flattening: all-pairs association on the plane must grow with L (constraints 1 and 2), and positional information must be compressed into scalar indices (constraint 3; for an early mitigation see segment-level recurrence [36]). A natural question follows: if we organize an additional axis of processing passes above the time axis—allowing the same content to be drafted first and revisited later—can we obtain, at the price of O(S) training and O(1) decode state, functions that the two-dimensional plane must in principle give up under causal constraints (Fig. 1, 2-D plane vs. 3-D computational lattice)?

### 1.2 Existing alternatives optimize "how to generate," not "what happens after generation"

Against constraints (1) and (2), four families of alternatives have emerged in recent years: state-space models (S4 [9], Mamba [10], Mamba-2 [11]), linearized attention [16,17,18,19], long convolutions (Hyena [15]), and the recurrence revival (RWKV [12], xLSTM [14], RetNet [13], descending from the RNN/LSTM lineage [3]); a 2025 Nature Machine Intelligence review summarizes this trend as "back to recurrent processing" [68].

We observe a common property of these works (offered here as an empirical observation, not a theoretical claim): all of them optimize the single function of "how to generate," and none introduces a second function beyond generation. A unidirectional causal structure cannot in principle revisit and revise what it has generated; prior work has observed the same phenomenon from the failure-mode side—the "articulation pathway" and the "execution pathway" of a model are structurally separate [30] (that work is still under review and is cited here only as an auxiliary observation). This paper tests the corresponding architectural hypothesis: explicitly differentiating the two functions of "drafting" and "revising" yields bidirectional revision ability without giving up causality—and at a cost far smaller than adding another layer of attention.

### 1.3 Why functional differentiation is necessary: a principled impossibility and an empirical exclusion

Two layers of evidence motivate a second function. (1) Principled: under causal constraints, the output at position p may not depend on information after p, so "rewriting the p-th token with both-sided context" has no legal computational path in a single-organ causal architecture—this is not a training-trick problem but a topology-of-information-flow problem. (2) Empirical (d128 tier): at the same parameter scale, a purely causal generative structure and a "generate + bidirectional refine" structure exhibit a 0-vs-1 capability exclusion on a synthetic infilling task (0.064 vs. 1.00)—the two functions are hard to carry simultaneously by a single structure. Together these elevate functional differentiation from a design preference to a structural necessity.

### 1.4 The proposed scheme and an operational definition of "organ"

FON is organized by functional division of labor (Fig. 1): the drafting organ replaces attention with multi-tap learnable delays and content-adaptive recursion, staying causal with O(S) training complexity and O(1) decode state; the revision organ keeps re-processing the written prefix with multi-scale ripple kernels, aided by gamma content addressing activated only at hole positions, and injects into the drafting organ's residual stream through a weak gate at every layer—never generating tokens. It is a reviser, not a second writer.

Operational definition of terms (avoiding metaphor-first reasoning): in this paper, a parameter subset is called a functional organ if it satisfies three criteria—(i) an independent parameter block; (ii) independently switchable by a structural signal (e.g., a mask); (iii) switching it off selectively impairs one ability while leaving the rest of language modeling unaffected (perplexity unchanged within measurement precision). The revision organ satisfies all three (§4.4, §4.5); note that criterion (iii) does not preclude the recovery of other abilities after switch-off—the recovery of retrieval after changing the coupling mode in §4.5 (0.23→1.00) is precisely mirror evidence of the coupling cost, consistent with the organ definition.

### 1.5 Contributions

1. **Architecture**: a drafting/revising dual-organ architecture inside a single language model, trained end-to-end with the pure LM loss, whose decode path contains no sequence-level quadratic attention (§3). 2. **Main performance**: better than Transformer (−33.7%) and Mamba2 (−14.1%) under iso-parameter, iso-token budget, three-seed mean; the advantage persists under both depth-aligned baseline classes (deeper TF-L8/L10 and weight-shared-recurrence TF/MM2-Looped) and holds at the 70M-parameter tier (T2 −22.8% / T1 −20.8%), narrowing but never reversing with scale (§4.2). 3. **Structural dividends**: only ~8–9% degradation at 8× extrapolation without positional-encoding engineering, with extrapolation stability between ALiBi and RoPE; O(1) decode state (§4.3). 4. **Bidirectional revision channel**: a structural pathway for the in-domain infilling ability that causal architectures lack in principle (+19% over on-demand coupling, +24% over drafting-only), with perplexity impact indistinguishable within measurement precision; out-of-domain (code) infilling is weaker than a same-recipe causal Transformer (§4.4). 5. **A complete cost-benefit ledger with mechanistic attribution**: a verdict experiment attributing long-training gains to recursion thickness rather than to the revision organ; a 2×2 attribution of the position kernel; three exclusionary pieces of evidence on the retrieval impairment; and a fill-mode battery control showing the synthetic battery is a dual-mode test the causal Transformer fails (FON is the only architecture here unifying generation and fill), with full reporting of the residual retrieval impairment and the wall-clock gap (§4.5, §4.6, §5).

---

## 2 Related work

### 2.1 Attention-free sequence modeling

State-space models obtain linear complexity through structured state transitions: S4 [9] established polynomial-coefficient state representations, Mamba [10] introduced input-dependent selectivity, and Mamba-2 [11] unified these under the SSD framework. Linearized attention replaces softmax with kernel approximations or feature maps (Linear Attention [16], Performer [17], GLA [18], Gated DeltaNet [19]); long convolutions cover long range with implicit filters (Hyena [15]); the recurrent lineage revives RNNs with gated linear units (RetNet [13], RWKV [12], xLSTM [14], descending from the RNN/LSTM lineage [3]); for adjacent implementations see HGRN [20], Based [21], Griffin [22], Jamba [23]; for compute allocation and test-time computation see Mixture-of-Depths [27] and TTT [28]; for internal modifications of attention see Differential Transformer [29] and FANformer [26]. A 2025 Nature Machine Intelligence review summarizes this trend as "back to recurrent processing" [68]. This paper takes Mamba2 as the representative baseline of this family, and covers the adjacent design point of weight-shared recurrence with Looped variants of TF/MM2; second-generation linear attention (GLA [18], DeltaNet [19]) is not included in the comparison—§4.2 states this explicitly as a scope declaration rather than unfinished work. **Shared boundary: all are single-function (generation) architectures.**

### 2.2 Bidirectionality and infilling

BERT [37] obtains bidirectional representations via masked language modeling but gives up generation; GLM [38] and InCoder [39] bring blank-infilling objectives into generative models, for pretraining and code completion respectively; early masked/infilling generative work [40,42] belongs to the same line; diffusion-style language models (LLaDA [43]) realize bidirectional generation via iterative denoising. FON's bidirectionality does not rely on training tricks: at hole positions the revision organ's bidirectional kernels are structurally activated (answers never enter the input, zero leakage), while non-hole positions are strictly causal.

### 2.3 Content addressing and memory

External differentiable memory appears in NTM [44], DNC [45], and Memory Networks [46]; retrieval-augmented LMs include kNN-LM [47], RAG [48], and RETRO [49]; for analysis of retrieval mechanisms inside attention see [67]. FON internalizes content addressing into a 0.066M-parameter hole-activated module.

### 2.4 Dual-system coupling and self-revision

The Bicameral Model [24] couples two independently pretrained LMs with a task-supervised interface; FON's two organs live inside one weight stack, are trained end-to-end with the pure LM loss, and the revision organ never generates. ResonatorLM [25] is close in spirit to the drafting organ's learnable recursive poles but has no second organ. Inference-time self-revision with external loops is studied in [50,51,52]; output-level fusion in [53]; the computational-power argument for weight-shared recurrence appears in Looped Transformers [54]—which §4.2 turns into a control baseline (Table 2).

**Table 1. Positioning along functional axes**

| Architecture | Sequence-level quadratic attention | Decode state | Bidirectional revision | Second organ | Supervision |
|---|---|---|---|---|---|
| Transformer [1] | yes | O(T) | none | none | LM |
| Mamba2 [11] / RetNet [13] | no | O(1) | none | none | LM |
| BERT [37] | yes | —(non-generative) | masked reconstruction | none | MLM |
| InCoder/FIM [39,41] | yes | O(T) | simulated by data reordering | none | LM+FIM |
| Bicameral Model [24] | yes | O(T) | cross-model coupling | yes (independent LMs) | task-supervised |
| **FON (this work)** | **no** | **O(1)** | **endogenous structural channel** | **yes (in-stack organ)** | **pure LM** |

### 2.5 Cognitive-science echoes (not motivations)

The dual-organ division is structurally isomorphic to complementary learning systems theory [58,59], reentrant processing [56,57] and adaptive resonance theory [60]; global workspace theory [61] offers a contrasting frame and [55] an early metaphor; see also [69]. These are post-hoc echoes: FON's division follows from the structural evidence of §1.3.

---

## 3 Architecture

### 3.1 Design principles

Fig. 1 FON (Fusion-Organ Network) architecture

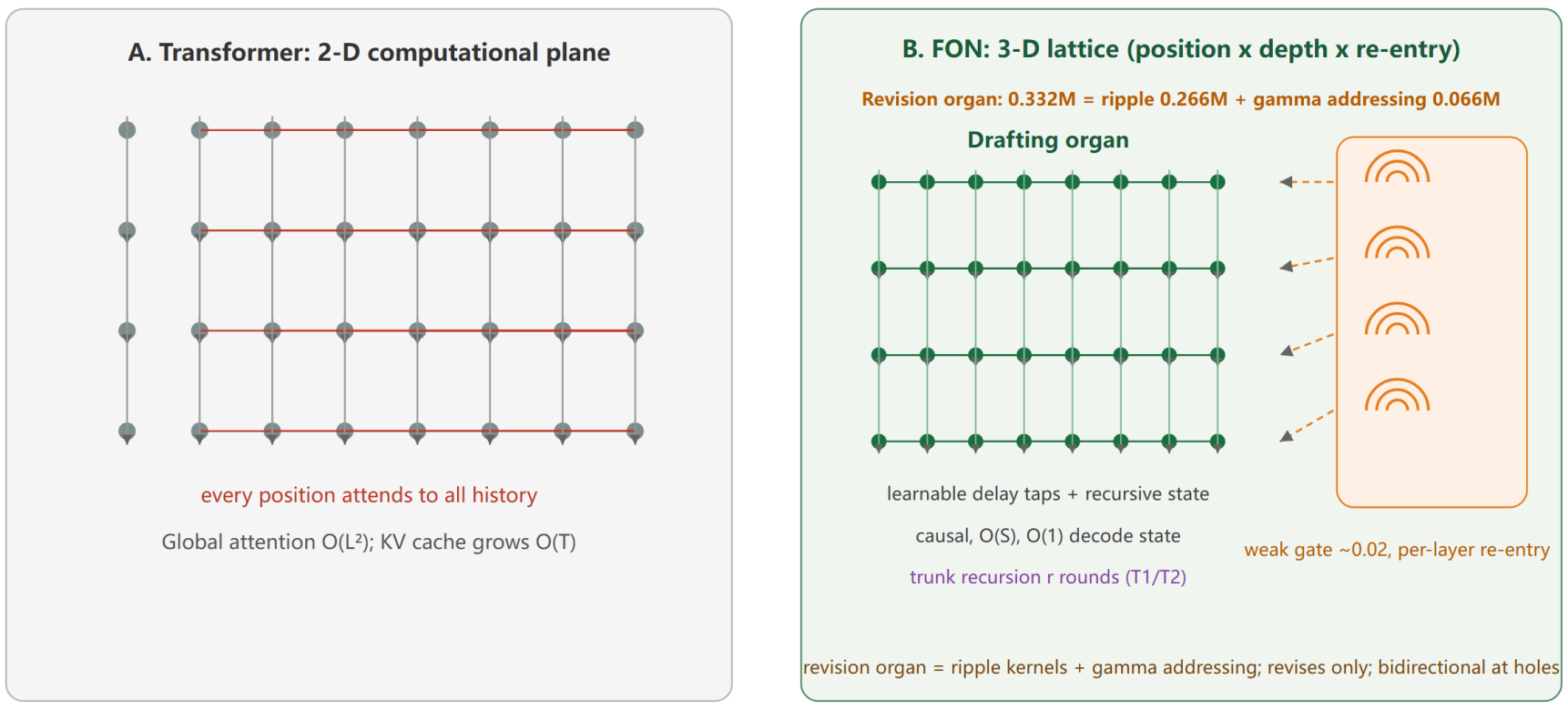

**Fig. 1.** FON architecture concept. Left: the Transformer's 2-D computational plane (all-pairs association, $O(L^2)$, KV cache $O(T)$); right: FON's 3-D computational lattice (drafting organ with local delay taps + recursion; revision organ's ripple tower re-enters every layer through a weak gate). Schematic.

(1) Causality is non-negotiable—outputs at non-hole positions may only depend on the prefix, a prerequisite for streaming deployment and generation consistency; (2) revision must be endogenous and leak-free—bidirectional processing is allowed only at hole positions where "the answer is not in the input," otherwise infilling ability could not be attributed to structure rather than data leakage; (3) fusion must be weak and frequent—revision signals are injected in small doses at every layer (rather than strongly at a single point), avoiding overwriting the drafting organ's converged representations. Each of the three principles has corresponding ablation evidence later (§4.6, §4.4). Fig. 2 gives a module-by-module data-flow comparison between FON and Transformer, and Fig. 2a and Fig. 2b expand the two stacks separately in the same visual language (reference Transformer architecture; FON architecture with the drafting-organ and revision-organ internals).

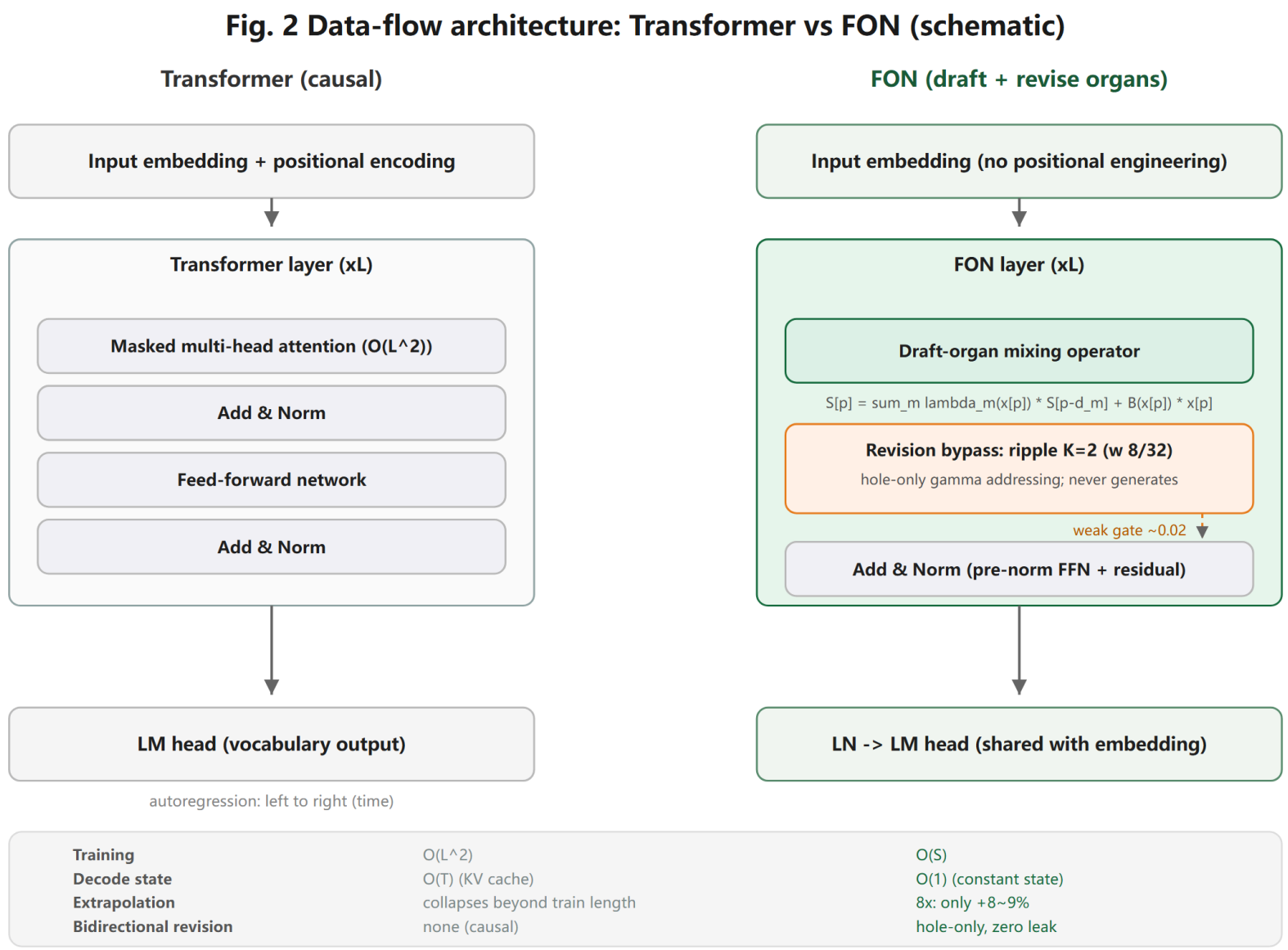

**Fig. 2.** Data-flow architecture comparison (schematic; Fig. 2a is the reference Transformer architecture): Transformer (embedding + positional encoding → L×[multi-head self-attention → Add&Norm → FFN → Add&Norm] → LM head) vs. FON (embedding without positional engineering → L×[drafting-organ mixing operator + revision-organ bypass with weak gate ≈0.02] → shared LM head); bottom comparison band: training $O(L^2)$ vs. $O(S)$, decode state $O(T)$ vs. $O(1)$, extrapolation collapse vs. only 8–9% degradation at 8×, bidirectional revision none vs. hole-legal only.

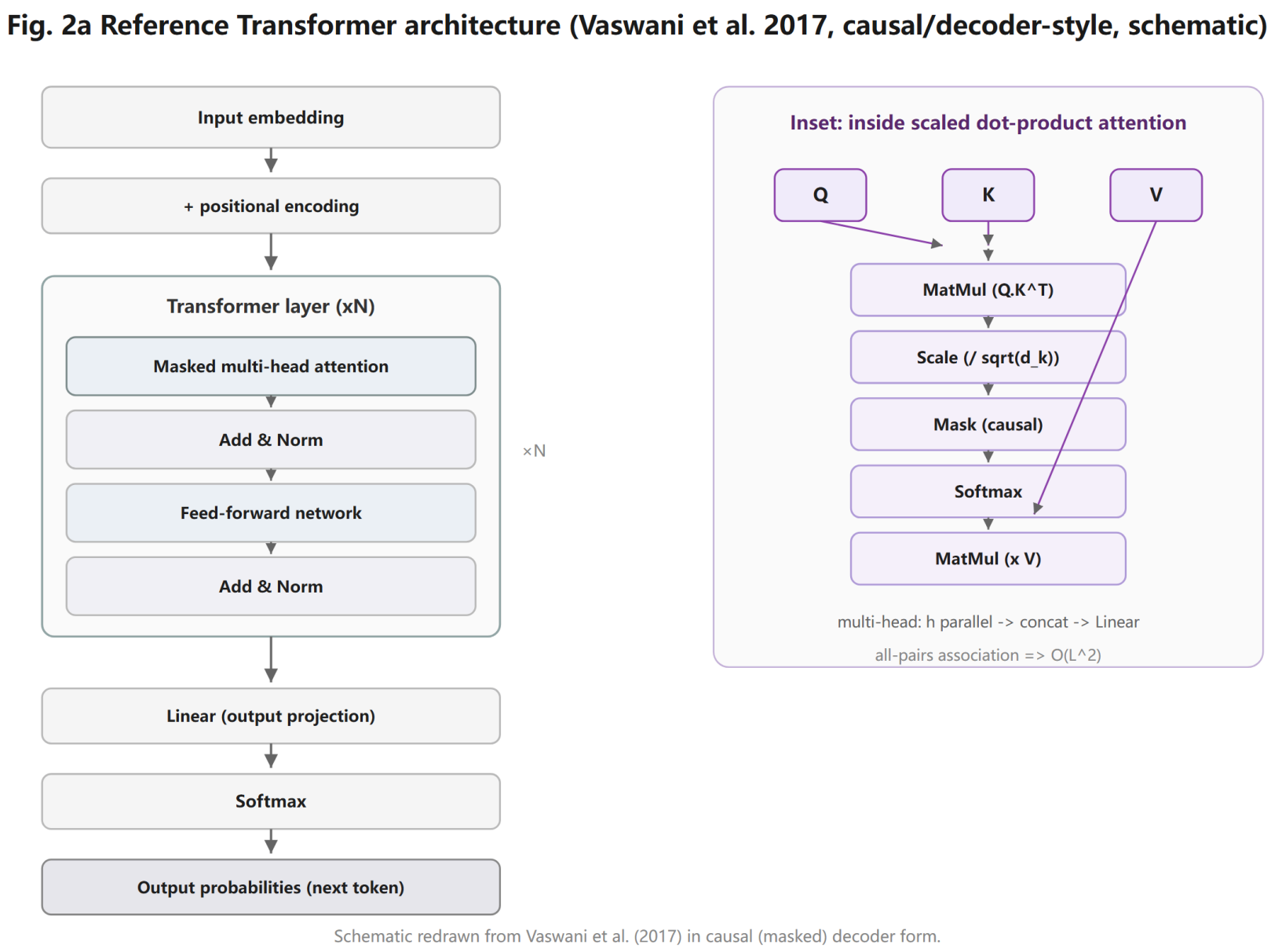

**Fig. 2a.** Reference Transformer architecture (Vaswani et al. 2017, redrawn in causal/decoder form): main pipeline plus an inset of scaled dot-product attention internals.

**Fig. 2b FON architecture (drafting organ + revision organ, schematic)**

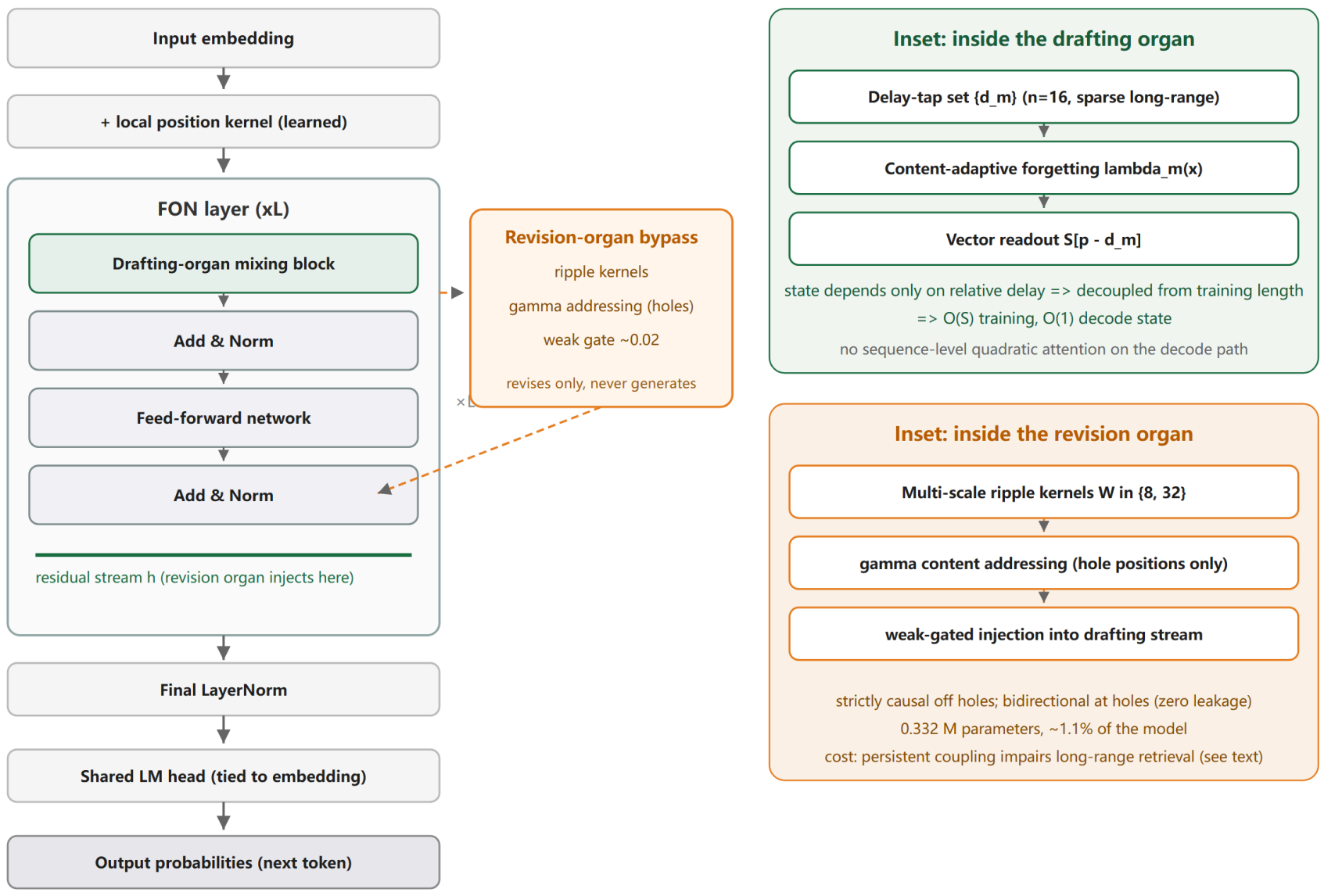

Schematic of the FON stack: drafting organ = multi-tap delays + content-adaptive recursion; revision organ = ripple kernels + gamma addressing.
Same visual language as Fig. 2a, for module-by-module comparison. T1 (r=1) / T2 (r=2) vary only the number of weight-reused passes of the drafting trunk.

> **Fig. 2b.** FON architecture (as implemented), drawn in the same visual language as Fig. 2a so that the two stacks can be compared module by module. Main pipeline: input embedding + learnable local position kernel → L × [drafting-organ mixing block → Add & Norm → FFN → Add & Norm] → final LayerNorm → shared LM head (tied to the embedding); a revision-organ bypass (multi-scale ripple kernels + gamma content addressing, injected through a weak gate of ≈0.02, revises only and never generates) feeds the residual stream at every layer. Inset (top right), inside the drafting organ: delay-tap set {d_m} (n=16, sparse long-range), content-adaptive forgetting λ_m(x), vector readout S[p−d_m]; because the state depends only on relative delay it is decoupled from the training length, giving O(S) training and an O(1) decode state with no sequence-level quadratic attention on the decode path. Inset (bottom right), inside the revision organ: ripple kernels W ∈ {8, 32}, gamma addressing active at hole positions only, weak-gated injection; strictly causal at non-hole positions and bidirectional at holes with zero leakage; 0.332M parameters (~1.1% of the model). Recursion thickness T1 (r=1) / T2 (r=2) varies only the number of weight-reused passes of the drafting trunk, not the structure.

### 3.2 The drafting organ: multi-tap delays + content-adaptive recursion

At position p:

$$S[p] = \Sigma_m \lambda_m(x[p]) \cdot S[p-d_m] + B(x[p]) \cdot x[p], \qquad y[p] = C(x[p]) \cdot S[p]$$

$\{d_m\}$ is a learnable delay-tap set (n=16, sparsely covering long range); $\lambda_m(x) = \exp(-\text{softplus}(\rho_m + W_\lambda x[p]))$ is a content-adaptive forgetting rate (selective, isomorphic to selective SSMs [11]); vector readout. All states depend only on relative delays—the structural source of extrapolation ability (§4.3). The position kernel provides a local positional prior and is a necessary component of performance at the d256 tier (§4.6).

### 3.3 The revision organ: ripple kernels + hole-activated gamma content addressing

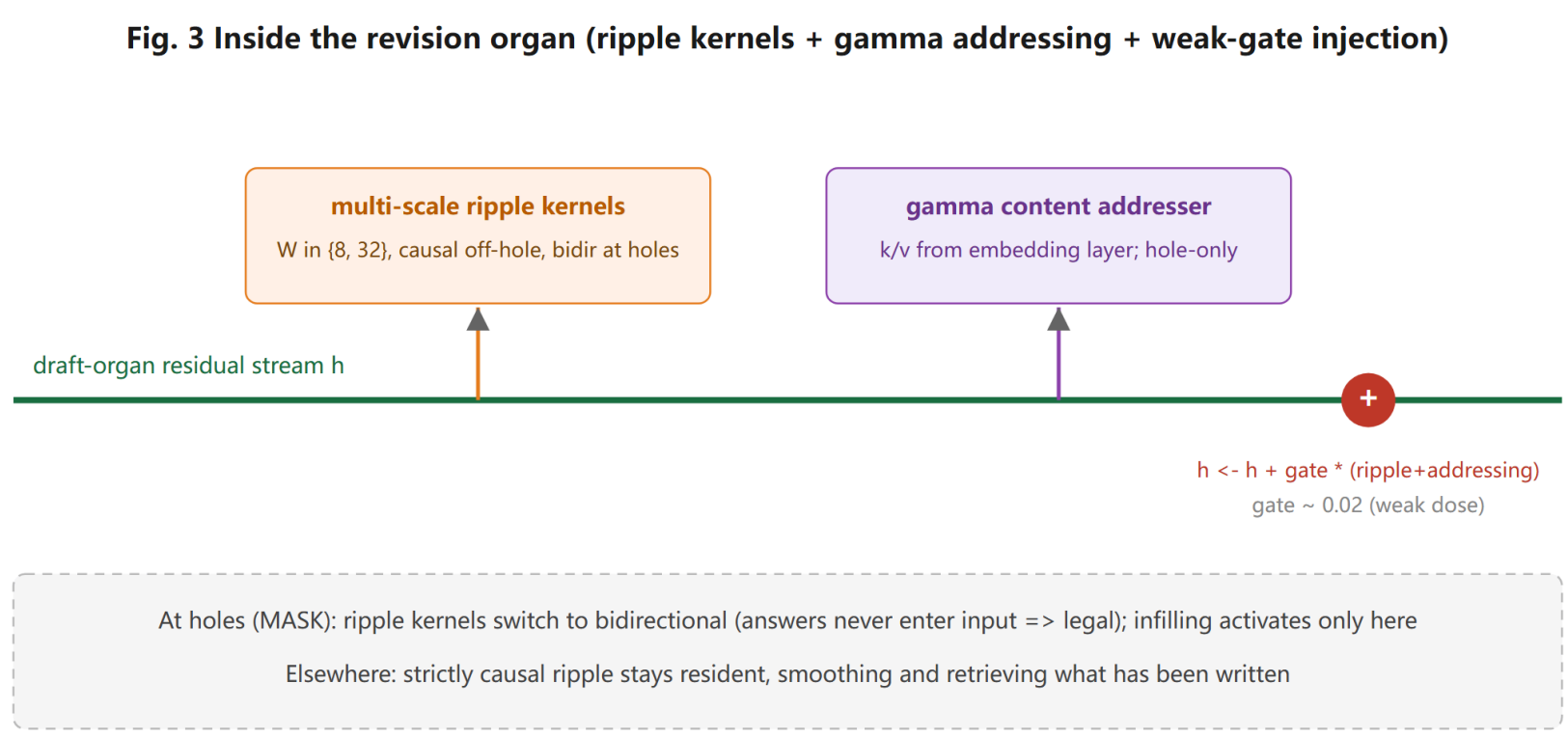

> **Fig. 3.** Inside the revision organ: multi-scale ripple kernels (W∈{8,32}, causal at non-hole positions, bidirectional at holes) + gamma content addressing (hole-activated, k/v from the token-embedding layer) + weakly gated injection. Schematic.

A K=2-layer ripple stack (0.266M parameters at the d256 tier) applies multi-scale ripples to the hidden-state sequence h (Fig. 3). Together with gamma addressing, the revision organ totals 0.332M parameters, ~1.1% of the model:

$$ripple(h)[p] = \Sigma\ (s \in \{8,32\})\ w_s \cdot Conv_s(LN(h))[p]$$

- At non-hole positions: kernels are strictly causal, zero leakage (bit-exact verification, Δ=0);
  At hole positions: kernels switch to bidirectional (legal: the masked answer never enters the input).

Gamma content addressing (0.066M) is activated only at hole positions: $inject_p = \gamma \cdot W_o(ctx_p)$, strictly zero at non-hole positions (code-level guarantee); keys/values come from the token-embedding layer rather than prefix hidden states, introducing no sequence-level state storage, so the decode state stays O(1). The addressing mechanism carries long-range exact retrieval (d128 lambda probe: mem_s1 = 0.86 at λ≥0.5 vs. 0.00 at λ=0 [d128·not verified at d256]); note that this probe shows the mechanism is capable of retrieval under ideal conditions, whereas synthetic long-range retrieval is still impaired under persistently coupled training—attribution in §4.5.

### 3.4 Weakly gated fusion

Each time the drafting organ completes a layer, the revision organ's K-layer ripple stack is invoked once in full (K×L re-entries per sequence):

$$h \leftarrow h + gate \odot mix(ripple(h),\ \gamma{\cdot}addr(h))$$

Dose sweeps show that a weak dose is a hard constraint [d128·not verified at d256]: gate=0.5 collapses entirely; the optimal plateau is 0.005–0.02; all layers share the same dose. During training the gate spontaneously converges to ≈0.015—the weak dose is a self-selected steady state of the network.

### 3.5 The thickness axis and three coupling modes

The drafting trunk recurs with weight sharing for r rounds (T1: r=1; T2: r=2, effective depth 8 layers). Recursion adds no parameters; total FLOPs rise from 1.07× to 1.35× Transformer (mixed-stack accounting in §4.2). **The main model of this paper is T1 (the iso-FLOPs fairness anchor); every** **piece of mechanistic evidence below is annotated with its accounting convention and scale tier.** Coupling modes: FON-AR (drafting-only) / FON-OD (on-demand coupling) / FON-PC (persistent coupling, the main model).

---

# 4 Experiments

## 4.1 Setup and fairness criteria

**Data and training**: WikiText-103 [8] (vocabulary 50,304); **the formal convention mixes in synthetic skill tasks at p_synth=0.35—the ratio was fixed early in project development, a conservative evaluation choice (the fill-mode battery is read against a causal-mode Transformer control in Table 8, where the causal baseline trails FON-PC on all six tasks)**; AdamW (β=(0.9, 0.95), wd 0.1, warmup 300, cosine, clip 1.0), lr 3e-4, batch 16, 20k steps, fp32; two phases: hole-free pretraining → hole-finetuning for 4k steps (lr 1e-4, p_fill=0.5, stride-8 p=0.15). **All models share one set of hyperparameters**—chosen during FON development; any bias favoring FON was not removed. Under this shared set, TF/MM2 perform at the level reported in the literature for this tier of WikiText-103 models; independent hyperparameter search for baselines is future work.

**Training-budget declaration (undertrained regime)**: 20k×16×256 ≈ 81.9M tokens ÷ 30M parameters ≈ 2.7 tokens/param, far below compute-optimal (~20 [5]). Conclusions are limited to the undertrained regime.

**Fairness quadruplet**: (1) iso-parameter for the main comparisons (28.9–31.0M; depth-aligned baselines relaxed to 34.2–35.8M to exclude the "depth" confound); (2) iso-FLOPs measured under two conventions (total plus mixed-stack excluding the shared vocabulary head, §4.2); (3) three seeds (seed 0/1/2), d256 noise band ±2.34%; throughout, σ denotes the population standard deviation across seeds (divisor n), which for n=3 is ~18% smaller than the sample convention (divisor n−1); (4) infilling is reported with protocol + baseline + net gain.

**Research questions**: RQ1 main performance; RQ2 structural dividends; RQ3 the bidirectional revision channel; RQ4 the revision organ's mechanistic role and cost; RQ5 design degrees of freedom. **Tier-labeling principle**: all d128-tier evidence is marked [d128·not verified at d256].

## 4.2 RQ1: Main performance

Main results appear in Table 2 and Fig. 4.

**Table 2. Main table (d256 L4, 20k steps, p_synth 0.35 convention)**

| Model | Params | Total FLOPs/token (×TF) | Mixed-stack FLOPs (×TF) | PPL (n) |
|---|---|---|---|---|
| Transformer [1] | 30.99M | 33.07M (1.00×) | 7.31M (1.00×) | 197.73 |
| Mamba2 [11] | 28.89M | 34.13M (1.03×) | — (1.03× not subdivided) | 152.53 |
| Transformer L8 (depth-aligned) | 34.17M | 40.38M (1.22×) | 14.62M (2.00×) | 159.58 (n=1) |
| Transformer L10 (depth-aligned) | 35.75M | 44.04M (1.33×) | 18.28M (2.50×) | 149.08 (n=1) |
| TF-Looped r2 (L4 weights ×2) | 31.01M | ≈46.1M (≈1.39×) | ≈46.1M (≈1.39×) | 182.73 (n=1) |
| TF-Looped r2 L6 | 32.59M | ≈48.4M (≈1.46×) | ≈48.4M (≈1.46×) | 163.52 (n=1) |
| MM2-Looped r2 | 29.71M | ≈68.3M (≈2.06×) | ≈68.3M (≈2.06×) | 139.43 (n=1) |
| FON-OD | 30.73M | 33.79M (1.02×) | 8.03M (1.10×) | 132.06 (n=3) |
| **FON-PC-T1 (main)** | 30.52M | 35.34M (1.07×) | 9.58M (1.31×) | **131.03 (n=3, σ 0.65%)** |
| FON-AR-T2 (drafting-only control) | 30.73M | 41.85M (1.27×) | 16.09M (2.20×) | 127.54 (n=3) |
| FON-PC-T2 | 30.52M | 44.64M (1.35×) | 18.88M (2.58×) | **126.75 (n=3, σ 0.79%)** |

**Two-convention declaration**: total FLOPs are 1.07× (T1) / 1.35× (T2) of Transformer; after excluding the vocabulary embedding and output head shared by all models (78% of total FLOPs, $2 \cdot V \cdot d \approx 25.76$M), the sequence-mixing stack itself is **1.31× (T1) / 2.58× (T2)**. We do not claim the architecture itself is iso-compute.

**What the last-row comparison means**: the drafting-only organ at equal recursion thickness already approaches FON-PC-T2 (127.54 vs. 126.75, a 0.6% difference, within the noise band; the equivalence boundary of equivalence claims is the noise band)—the revision organ's perplexity contribution is indistinguishable within measurement precision. Rather than weakening the paper, this delimits its positioning: the revision organ's value is not in language-modeling scores (§4.4, §4.5).

**Fairness-risk self-declaration (with depth-aligned verdicts)**. The main comparison Transformer is L4, while FON-T2 has effective depth 8 (weight-shared). We therefore completed depth-aligned baselines and declared the verdict criteria in advance:

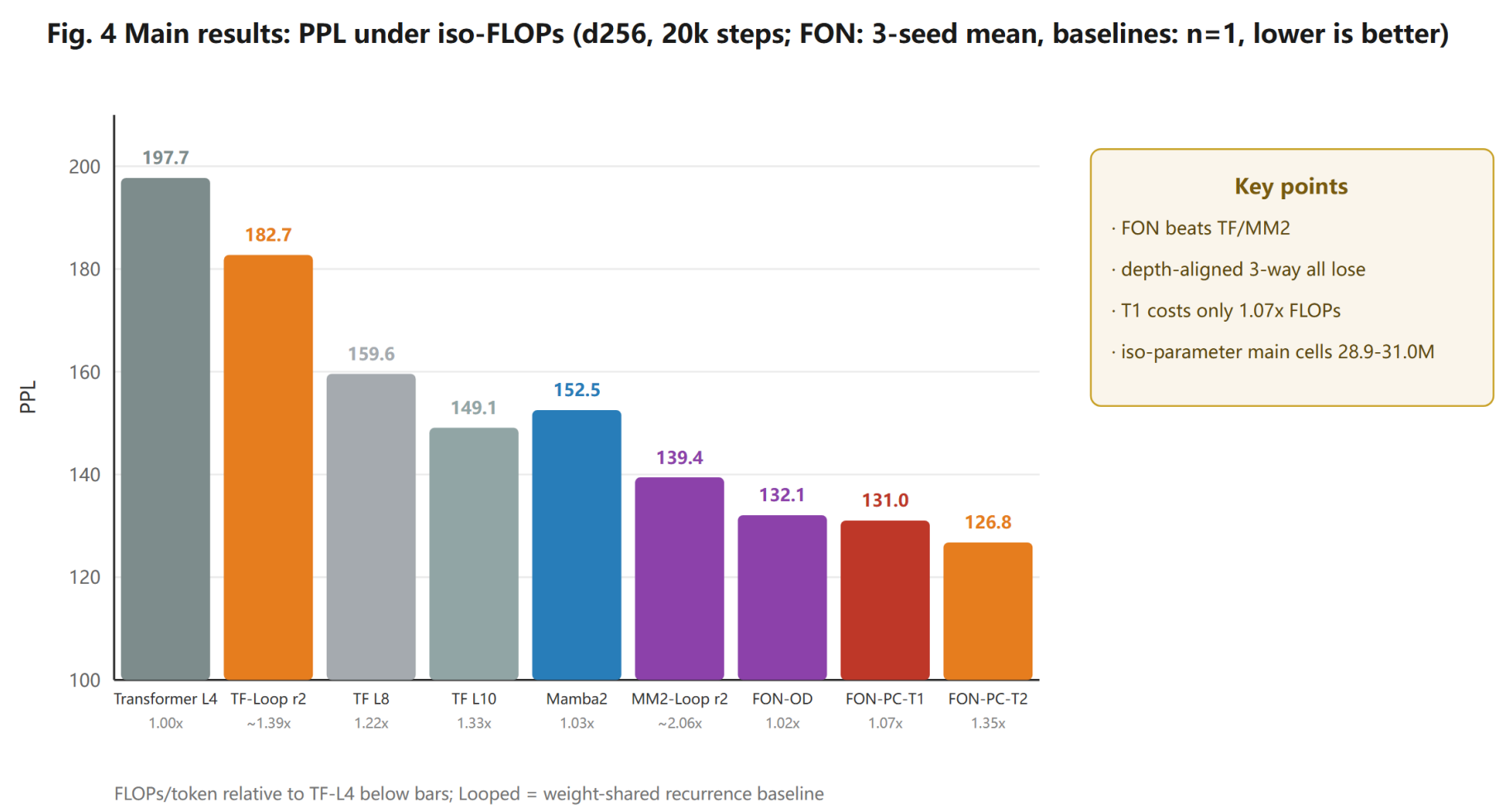

**Fig. 4.** Main results (d256 L4, 20k steps, p_synth 0.35). Bars = PPL (FON/OD are 3-seed means, σ ≤ 0.79%; baselines n=1), with total FLOPs/token relative to Transformer below bars. Includes three depth-aligned baselines: TF-L8/L10 (independently deepened) + TF/MM2-Looped-r2 (weight-shared recurrence). Across all depth-aligned configurations, FON-PC-T1 wins consistently (19.9%~33.7%).

- **Transformer L8 / L10 (approximately iso-parameter, 34.17M/35.75M) reach 159.58 / 149.08**. Verdict: (i) at equal effective depth (TF-L8 vs. FON-PC-T2, both 8-layer-equivalent) FON is better by **20.6%**; (ii) at equal mixed-stack compute (TF-L10 2.50× vs. FON-PC-T2 2.58×) FON is better by **15.0%**; (iii) even deepening the Transformer to 10 layers (+15–17% parameters over the main comparison, 2.50× mixed-stack compute) still falls short of the cheapest FON-PC-T1 (1.31× mixed-stack). **The "advantage comes from depth" falsification fails.**

**Looped Transformer / Mamba2 (weight-shared recurrence)**: TF-Looped-r2 (31.01M, effective depth 8) = **182.73**, only 7.6% better than TF-L4; TF-Looped-r2-L6 (32.59M, effective depth 12) = 163.52; MM2-Looped-r2 (29.71M, effective depth 8) = 139.43. **Verdict: the "advantage comes from recursive depth / weight sharing" falsification fails on all five configurations**—FON-PC-T1 (effective depth only 4) is still 19.9% better than the deepest Looped variant (effective depth 12). The advantage comes from the delayed-mixing operator itself, not from depth.**Scope declaration**: second-generation linear-attention baselines of the same family (GLA / DeltaNet) are not included: this paper already includes Mamba2 (representative of modern recurrent memory) and Looped (the strongest "depth-aligned" falsification); delta-rule baselines add little discriminative power and are costly to implement fairly, and are explicitly declared out of scope rather than pending.Pure-LM sub-table (three-seed final values):

**Table 2b. Pure-LM convention (p_synth=0, d256, 20k steps)**

| Model | PPL (3-seed mean ± σ, per-seed values) |
|---|---|
| Transformer | 150.05 (151.97 / 146.92 / 151.27, σ 1.5%) |
| Mamba2 | 115.60 (116.69 / 114.27 / 115.83, σ 0.86%) |
| **FON-PC-T1** | **106.30 ± 0.72** (107.31 / 105.66 / 105.94, σ 0.68%) |
| FON-PC-T2 | 102.57 (n=1) |

Against the Transformer three-seed mean: T1 **−29.2%**, T2 −31.6%; against Mamba2: T1 −8.0%. Under the pure-LM convention (synthetic skills removed), TF benefits more (197.73→150.05), but FON's advantage holds in the same direction, and FON-PC-T1's seed variance under this convention (0.68%) remains below TF's (1.5%).

**Table 3. Scale consistency (d512, b16, 20k steps)**

| Model | PPL |
|---|---|
| Transformer | 118.46 |
| FON-AR-T2 (drafting-only control) | 92.0–92.5 (n=2) |
| FON-PC-T2 | **91.46 (n=3; 92.07 / 91.26 / 91.05, σ 0.48%)** |
| FON-PC-T1 | 93.83 (n=1) |

The advantage stays significant but relatively narrows with scale (d256 −35.9% vs. Transformer; d512 −22.8%, T1 convention −20.8%), never reversing; results are reported per tier without linear extrapolation [4]. The T1/T2 gap at d512 is +2.6%, consistent with the d256 tier (+2.6%; +3.46%/+2.56% under the matched-r conventions)—the recursion-thickness dividend persists with scale.

**Table 4. Efficiency audit (B8/S256/fp32 measurements, 10-09 compute-ledger convention)**

| Model | Total FLOPs/token (d256) | rel. TF | ms/step | Decode state |
|---|---|---|---|---|
| Transformer | 33.07M | 1.00× | 28.7 | O(T) |
| FON-PC-T1 | 35.34M | 1.07× | 186.0 (6.5×) | O(1) |
| FON-PC-T2 | 44.64M | 1.35× | 351.7 (12.3×) | O(1) |

**Engineering disclosure**: FON's current research implementation has a 6.5–12.3× wall-clock of Transformer. This is an implementation gap, not an architectural compute gap (FLOPs differ by only 1.07–1.35×), so all wall-clock comparisons are conservative for FON. Measured curves of per-token decode latency and peak memory versus context length (512→32k) are future work (if FON turns out slower for lack of kernel optimization, the curves will be drawn as-is and labeled an implementation gap).

### 4.3 RQ2: Structural dividends—extrapolation and decode state

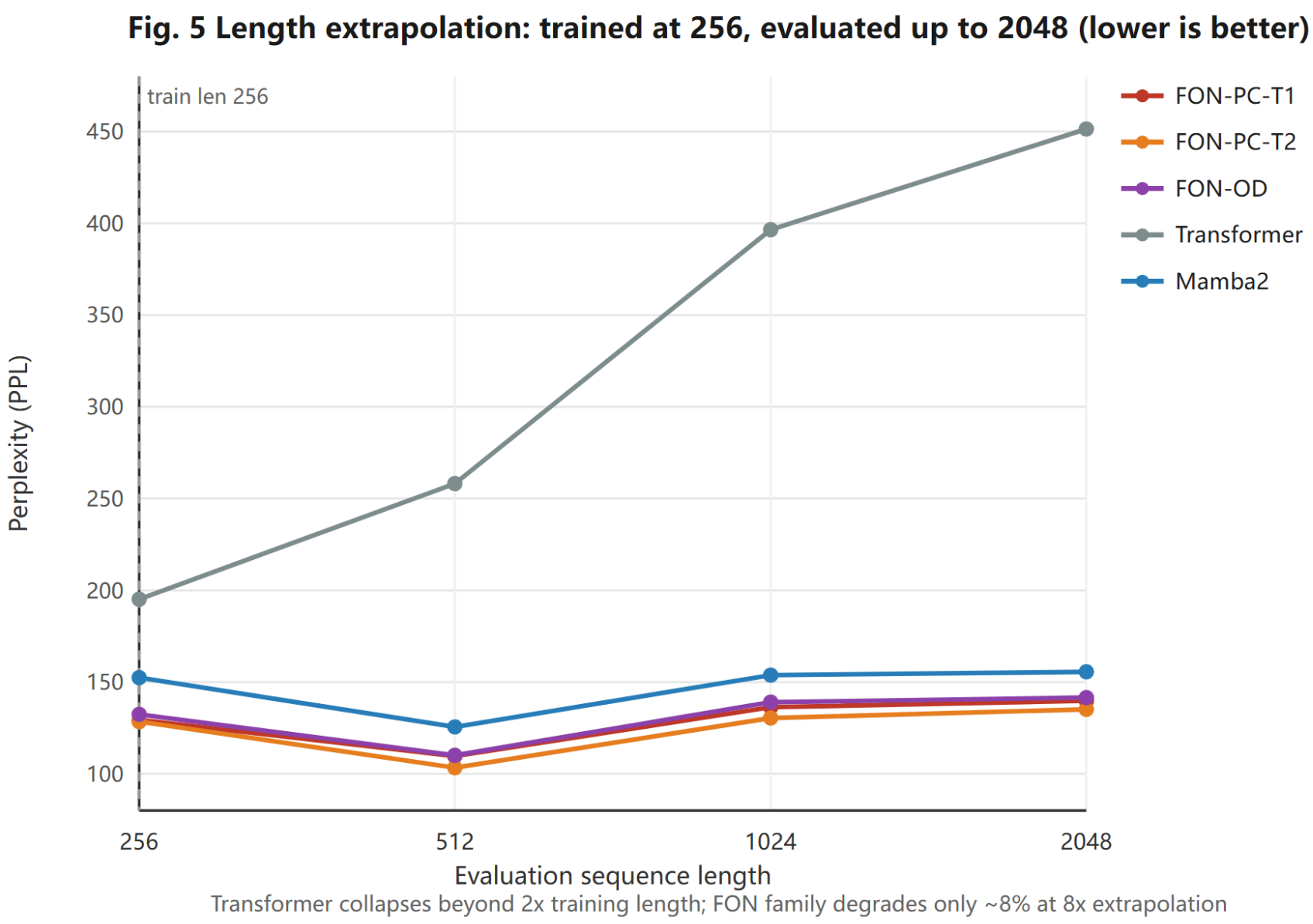

> **Fig. 5.** Length extrapolation (trained at 256, evaluated up to 2048, direct protocol). The FON family degrades ~8–9% at 8×; the Transformer (absolute PE) collapses after 2×. Positional-engineering controls (matched 8k-step budget): absolute PE +59.9% / RoPE +18.6% / ALiBi +2.6%—FON (+9.2%) reaches RoPE-tier stability without positional engineering (numbers in §4.3 text).

Trained at length 256 and evaluated directly up to 2048 (Fig. 5; the evaluation protocol follows RULER's long-context judgment principles [64]): the FON family degrades by only ~8–9% at 8× extrapolation (FON-PC-T1: 129.4→139.8; d512 T1 single point 92.9→101.5, +9.2%); the Transformer (learnable absolute positional encoding) collapses after 2× (195→451); Mamba2's curve is flat (152→156) but its absolute level trails FON throughout. **Positional-engineering controls (matched 8k-step budget, d256, relative degradation 256→2048)**: absolute PE +59.9% (318.45→509.27), RoPE +18.6% (241.87→286.96), ALiBi **+2.6%** (226.15→232.13). Hence the honest conclusion: **FON achieves RoPE-tier extrapolation stability without any positional-encoding engineering** (FON +9.2% vs. ALiBi +2.6%; ALiBi is smoother at the price of attenuating far-range attention throughout). Our claim is limited to "extrapolates without positional-encoding engineering" (relative-delay keys decouple positional signals from the training length), not "extrapolates better than every positional-engineering scheme." Re-running the extrapolation anchors under the main-table convention is future work (currently a 1.2% discrepancy from evaluation chunking).

Decode state (Fig. 6): FON keeps a fixed number of taps per layer (n=16), independent of sequence length (O(1)); the Transformer KV cache grows linearly with length (O(T)).

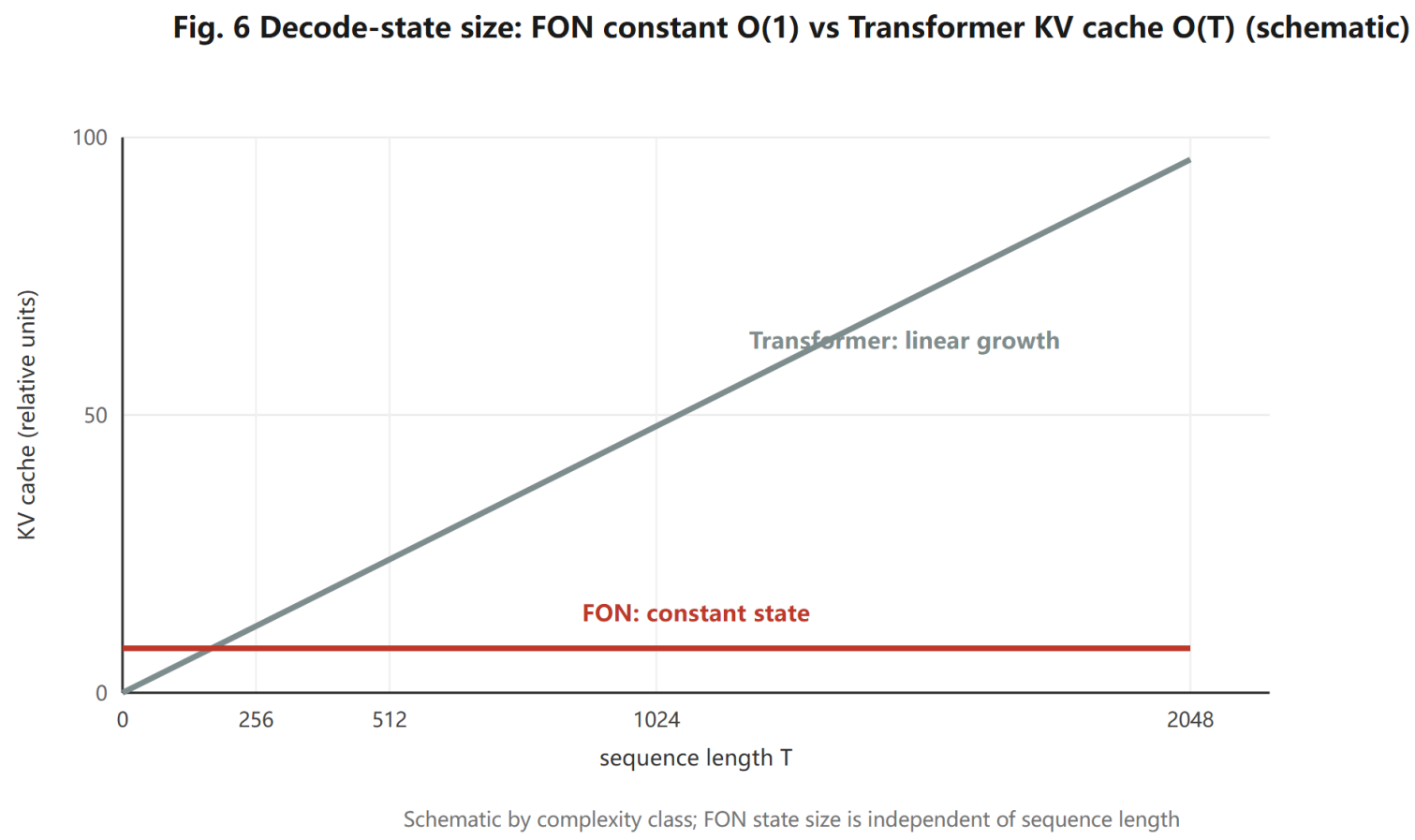

**Fig. 6.** Decode-state size (complexity class, schematic). FON O(1) vs. Transformer O(T).

### 4.4 RQ3: The bidirectional revision channel

Infilling results appear in Table 5 and Fig. 7. **Evaluation protocol (declared first)**: full-vocabulary argmax hit rate, chance = 1/50304 ≈ 0.002% (verified at code level); the key baseline is the same-architecture no-hole model—whose fill is nonzero because the LM prior itself can guess words.

**Table 5. Infilling ability (d256; PPL and fill come from different checkpoints—hole-free pretraining vs. ft4k, noted separately)**

| Model | checkpoint | no-hole baseline | after ft4k | Net gain |
|---|---|---|---|---|
| Causal TF (WikiText fill) | pretrain / ft4k | — (not measured) | 0.051 | prefix-only infilling is learnable (code domain: no-hole 0.032 → 0.227, see Note (4)) |
| FON-AR-T2 (drafting-only) | ft4k | ~0.19 | 0.194 | ≈0 (LM prior) |
| FON-OD | ft4k | — | 0.203 | — |
| FON-PC-T1 | ft4k | — | 0.233 | — |
| **FON-PC-T2** | ft4k | — | **0.241** | **+0.04~0.05 over drafting-only (d256)** |
| FON-PC-T2 (d512) | ft4k | — | **0.273** | **+0.07 over drafting-only ~0.20 (d512)** |

**Notes**: (1) Absolute fill levels are of the same order as the LM prior; fill is not a primary metric of this paper. (2) The structural fact is that **persistent coupling is stably above the drafting-only organ** (d256 net +0.04~0.05; d512 +0.07; +19% over on-demand coupling and +24% over drafting-only at d256; +35% over drafting-only at d512). On whether causal architectures can infill at all, the answer depends on the domain: the no-hole baseline for out-of-domain code FIM is 0.000 (measured for the whole FON family; the causal TF still shows a 0.032 LM-prior leakage), while the in-domain WikiText no-hole baseline is of LM-prior order (~0.194 for drafting-only, Table 5). The earlier statement that "causal architectures are structurally incapable of infilling" has been falsified by the TF-ft4k measurements of

v5.2 (0.051 in-domain, 0.227 out-of-domain) and is retracted: a causal architecture can learn **prefix-only** infilling through hole finetuning, but "bidirectional revision that uses the suffix" still has no legal computational path in a causal architecture (§1.3(1)). (3) The "hole tax" (the PPL cost introduced by hole finetuning, ft4k convention) is +10.1–11.0% for the FON variants and +7.3% for the causal Transformer at d256 (same-protocol per-checkpoint pairs: TF 197.73→212.08; FON-PC-T1 131.72→145.52; FON-PC-T2 127.69→141.13; FON-OD 131.32→145.77; FON-AR-T2 128.23→141.16). It is therefore large but not strictly architecture-independent—note this is defined differently from the d128 always-holed convention (+1.7%) and the two must not be mixed. (4) **Code-domain FIM (same-recipe controls)**: reference models finetuned with the same ft4k hole recipe are evaluated under the same protocol as FON. Results: FON-PC-T2 0.0445 > FON-OD 0.0418 > FON-PC-T1 0.0352 (FON no-hole 0.000, net gain +0.035~0.045); **causal Transformer 0.227 (no-hole 0.032, net gain +0.195); bidirectional Transformer 0.003**. Hence: FON's infilling advantage is limited to the in-domain setting (WikiText fill: FON 0.233 vs. same-recipe causal TF 0.051) and reverses to a disadvantage out of domain on code (0.045 vs. 0.227)—a plain causal Transformer exploits local copy patterns in code far better than FON's delayed-mixing operator. The bidirectional TF, despite bidirectional attention, fails at cross-domain infilling (0.003); its causal-protocol PPL ≈1.35 is a label-leakage artifact and is not meaningful. (5) Stronger mechanistic evidence (d128): the bidirectional-refinement structure and the causal-generation structure exhibit a 0-vs-1 capability exclusion (1.00 vs. 0.064); the narrowed claim here is—**a single weight stack can host both modes end-to-end under a pure LM loss: strictly causal at non-hole positions (verified bit-exact against a purely causal model) and bidirectional at hole positions (zero leakage, verified)**.

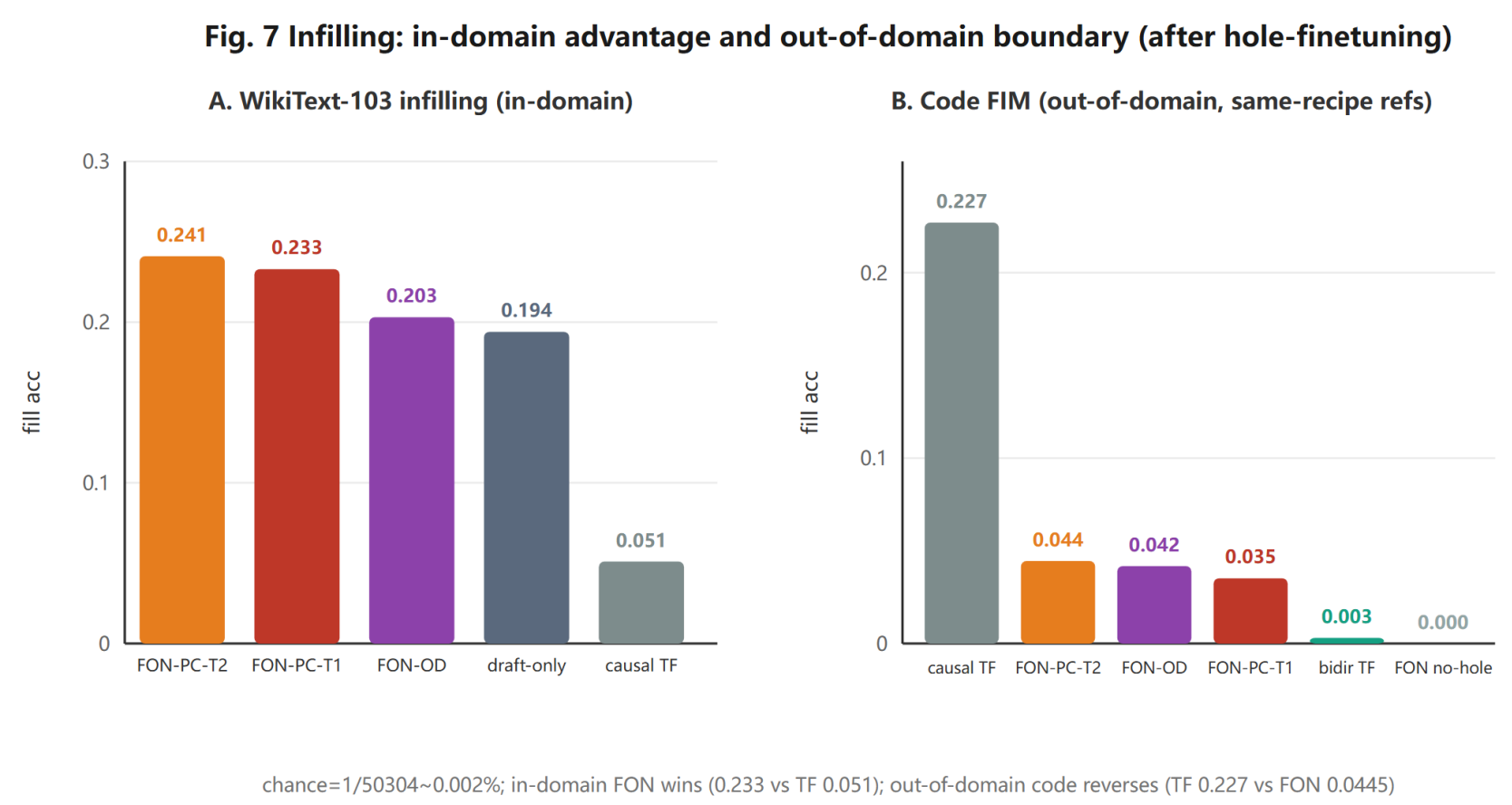

> **Fig. 7.** Bidirectional infilling (d256, after ft4k, full-vocabulary argmax, chance=1/50304). A: WikiText fill (drafting-only no-hole baseline 0.194; same-recipe causal TF reference 0.051); B: code FIM (FON no-hole=0.000; same-recipe causal TF reference 0.227, bidirectional TF reference 0.003—the out-of-domain infilling disadvantage is discussed in §4.4 Note (4)).

### 4.5 RQ4: The revision organ's mechanistic role and cost

**To the question "the revision organ's perplexity contribution is indistinguishable within measurement precision—why keep it,"** this paper's evidence chain answers: it does not improve the language-modeling score; it extends the capability envelope (the infilling relative advantage in Table 5, the 0-vs-1 exclusion on synthetic tasks)—at no measurable score cost.

**Attribution verdict for long-training marginal gains (d128, Fig. 8).** In the early comparison, on-demand coupling (T1) showed zero gain from 4k→8k (427.92→429.22) while persistent coupling (T2) showed −28.5% (420.84→301.08); but that comparison changed two variables at once—"revision switch" and "recursion thickness"—and could not be attributed. The verdict control (drafting-only T2 over the same interval) completes the picture:

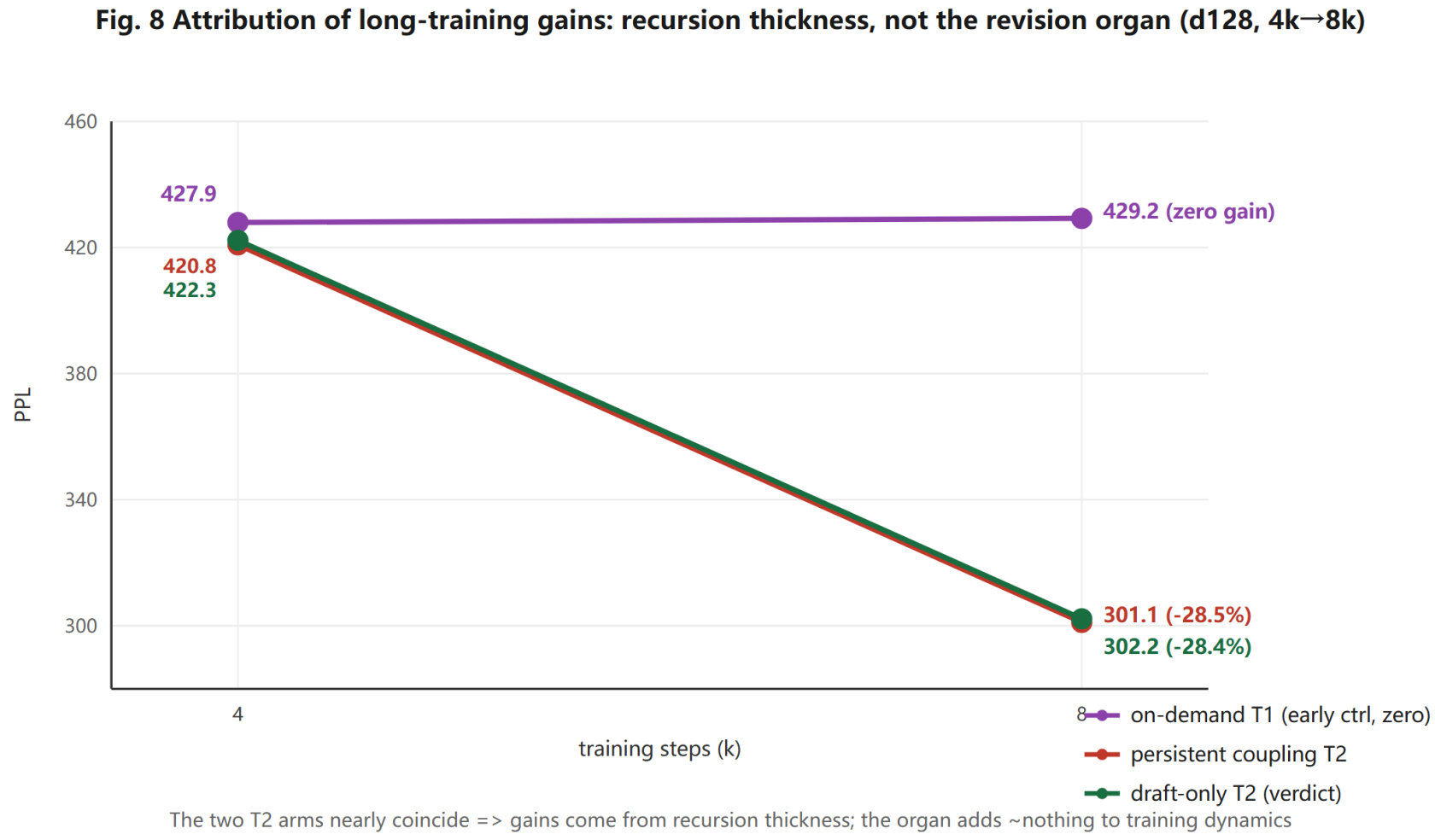

> **Fig. 8.** Attribution verdict for long-training marginal gains (d128, 4k→8k). On-demand coupling T1 (early control) shows zero gain; drafting-only T2 (verdict control) and persistent coupling T2 nearly coincide (−28.4% vs. −28.5%) ⇒ gains are explained by recursion thickness; the drafting-only-T1 arm was not run (the two T2 arms already isolate the revision variable).

**Table 6. Attribution of long-training marginal gains (d128, 4k→8k steps)**

| Arm | 4k | 8k | Marginal gain |
|---|---|---|---|
| On-demand coupling T1 (early control) | 427.92 | 429.22 | zero gain |
| Drafting-only T2 (verdict control) | 422.25 | 302.24 | **−28.4%** |
| Persistent coupling T2 (FON-PC precursor) | 420.84 | 301.08 | −28.5% |

Note: the drafting-only-T1 arm was not run (it does not affect the verdict: the two T2 arms share thickness and differ only in whether the revision channel is resident). The two T2 arms nearly coincide—**the −28% long-training gain is explained by recursion thickness, not by the revision organ**. Together with the FON-AR-T2 row of Table 2, the revision organ's contributions to both the language-modeling score and training dynamics are indistinguishable within measurement precision. This result contradicts an earlier hypothesis of ours (that the revision organ amplifies long-training dividends); we report this negative attribution as-is. The revision organ's justification converges to the capability envelope.

**Trade-offs along the coupling-strength axis (Fig. 9, Table 7).**

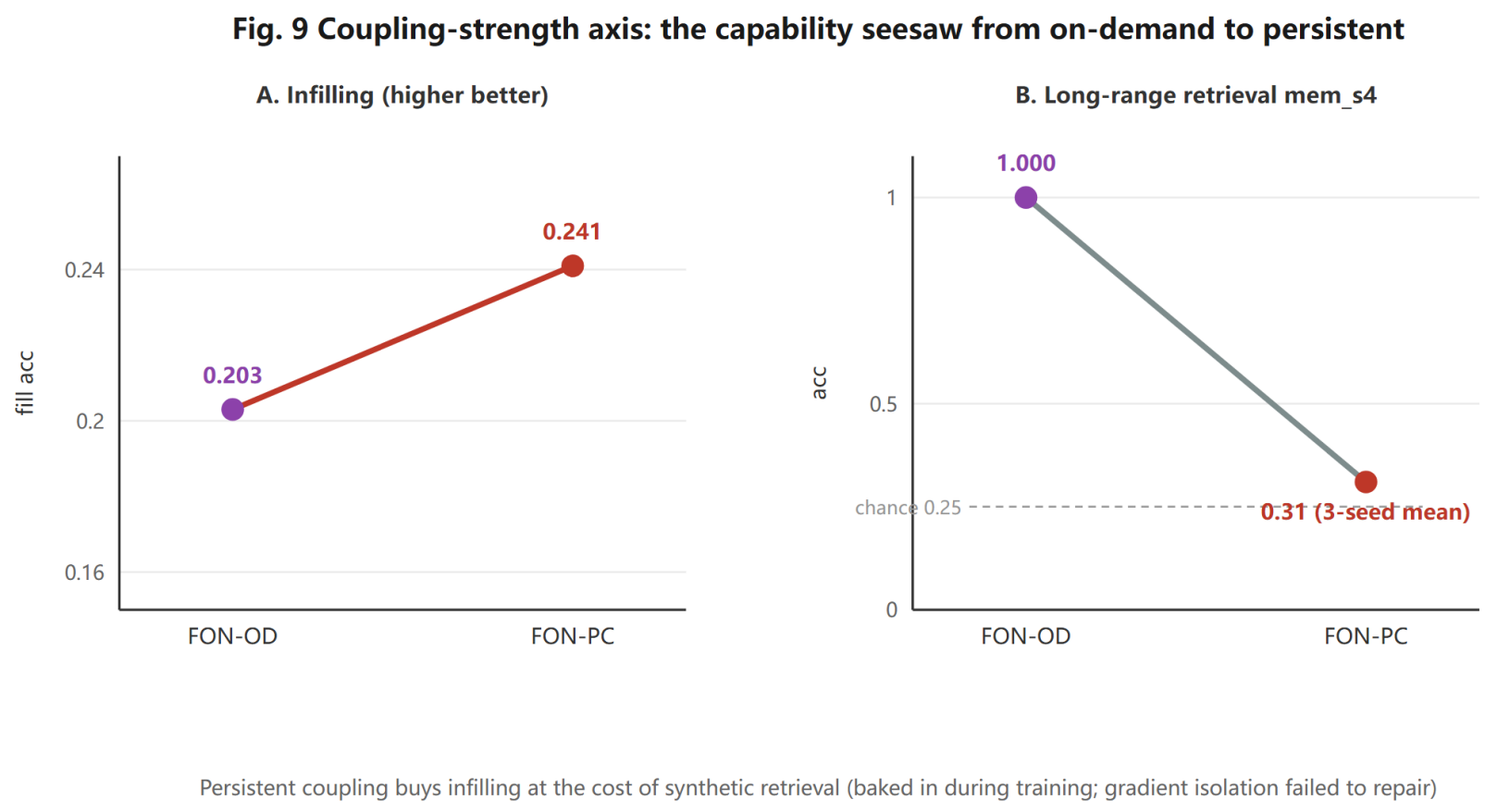

> **Fig. 9.** Capability seesaw along the coupling axis. A: infilling (OD 0.203 → PC 0.241); B: long-range retrieval mem_s4 (OD 1.000 → PC 0.31, 3-seed mean; chance = 0.25).

**Table 7. Coupling-strength axis (AR → OD → PC)**

| Metric | FON-AR | FON-OD | FON-PC | Reading |
|---|---|---|---|---|
| Causal PPL (d256, T2) | 127.54 | — | **126.75** | revision organ PPL contribution ≈ 0 |
| Infilling fill (d256, ft4k) | 0.194 | 0.203 | **0.241** | PC over OD +19% |
| Long-range retrieval mem_s4 (d512) | **1.000** | — | 0.23–0.26 | impaired by persistent coupling |
| Long-range retrieval mem_s4 (d256) | **1.000** (n=3, measured) | **1.000** | 0.28–0.37 | same |

**Attribution of the impairment (exclusionary evidence, not positive causation)**: (1) switching off the fusion gate at inference does not restore retrieval (0.272 vs. 0.280)—ruling out inference-time interference; (2) gradient isolation of hole batches (a controlled variant with bit-identical forward) does not repair it—ruling out hole-batch coupling; (3) Mamba2 collapses identically (0.263)—localizing it as a shared difficulty of architectures without global attention. These are three exclusionary pieces of evidence; positive causal evidence—training-time mem_s4 trajectories and the dose–retrieval Pareto front—is future work.

**Table 8. Task-battery comparison (d256 convention, fill-mode evaluation, reported in full)**

| Task | Transformer (causal) | Transformer (bidirectional, non-generative encoder) | Mamba2 | FON-AR (n=3) | FON-OD (n=3) | FON-PC (n=3) |
|---|---|---|---|---|---|---|
| mem_s4 (long-range retrieval) | 0.19 | **1.000** | 0.263 | **1.000** | **1.000** | 0.28–0.37 |
| mem_l (long-range memory) | 0.21 | **1.000** | 0.278 | **1.000** | **1.000** | 0.29–0.39 |
| add (arithmetic) | 0.31 | **0.927** | 0.480 | 0.52–0.62 | 0.59–0.68 | 0.47–0.63 |
| state_l (state tracking) | 0.37 | **1.000** | 0.709 | **1.000** | **1.000** | 0.96–0.99 |
| sort_rel (sorting) | 0.50 | **0.923** | 0.461 | 0.64–0.68 | 0.62–0.68 | 0.52–0.53 |
| cmp (comparison) | 0.78 | 0.946 | 0.794 | **0.94–0.96** | 0.94–0.95 | 0.94–0.95 |

All FON columns are per-seed minima–maxima over three seeds (d256, 20k steps, p_synth 0.35; FON-AR = FON-AR-T2, FON-PC = FON-PC-T2, matching the three-seed PPL convention of Table 2); Transformer and Mamba2 are n=1. **Evaluation mode (declared explicitly).** The battery is a fill-mode test: the answer occupies a MASK position and may draw on context on both sides. For attention this is full bidirectional self-attention (the "Transformer (bidirectional, non-generative encoder)" column); for FON it is the hole-activated bidirectional revision organ; Mamba2 cannot switch direction and stays causal. **"Transformer (causal)" is a mode control on the same checkpoint, re-evaluated with the causal mask on**—it reports the standard autoregressive Transformer's ceiling on these tasks. We retain the bidirectional column deliberately and report it transparently: it is **not** a generative competitor to FON but a three-in-one reference—(i) the **ceiling** for fill-mode capability under full bidirectional attention, (ii) a **directional ablation** isolating how much of the battery is attributable to bidirectionality rather than causality, and (iii) the **BERT-style non-generative encoder** baseline that no other row covers. Omitting it would misleadingly suggest FON "loses" on this battery, whereas the battery is a fill-mode test that the causal (autoregressive) Transformer fails outright.

The mode control overturns the naive reading (Fig. 12). Bidirectionally the Transformer tops the battery—but that is a **bidirectional, non-generative encoder** (BERT-style): by construction it cannot generate, so it is the fill-mode ceiling rather than a generative baseline. **Causally, the same checkpoint collapses and is outperformed by FON-PC on all six tasks** (mem_s4 0.19 vs. 0.28–0.37; add 0.31 vs. 0.47–0.63; state_l 0.37 vs. 0.96–0.99; cmp 0.78 vs. 0.94–0.95), even though the evidence sits in the left context. The battery therefore does not mark a FON weakness: it measures fill capability, which requires bidirectional processing that causal attention lacks. FON is the only architecture here that pairs causal generation with fill capability—FON-AR/OD reach the bidirectional Transformer's 1.000 on retrieval and state tracking while remaining causally generative; the bidirectional Transformer cannot generate and Mamba2 cannot switch direction. On the future-dependent fill tasks (ind_r, state_inv, perm_h1, mem_r, trans_r) the same split is sharper still: the causal Transformer sits at or below chance (e.g., ind_r 0.02, state_inv 0.09, perm_h1 0.09; single-seed) while FON variants reach 1.000. Note: FON-PC's state_l ≈ 0.96–0.99 coexisting with mem_s4 ≈ 0.28–0.37 is not contradictory—the former tests local state tracking (short range), the latter long-range content retrieval; the two rely on different mechanisms.

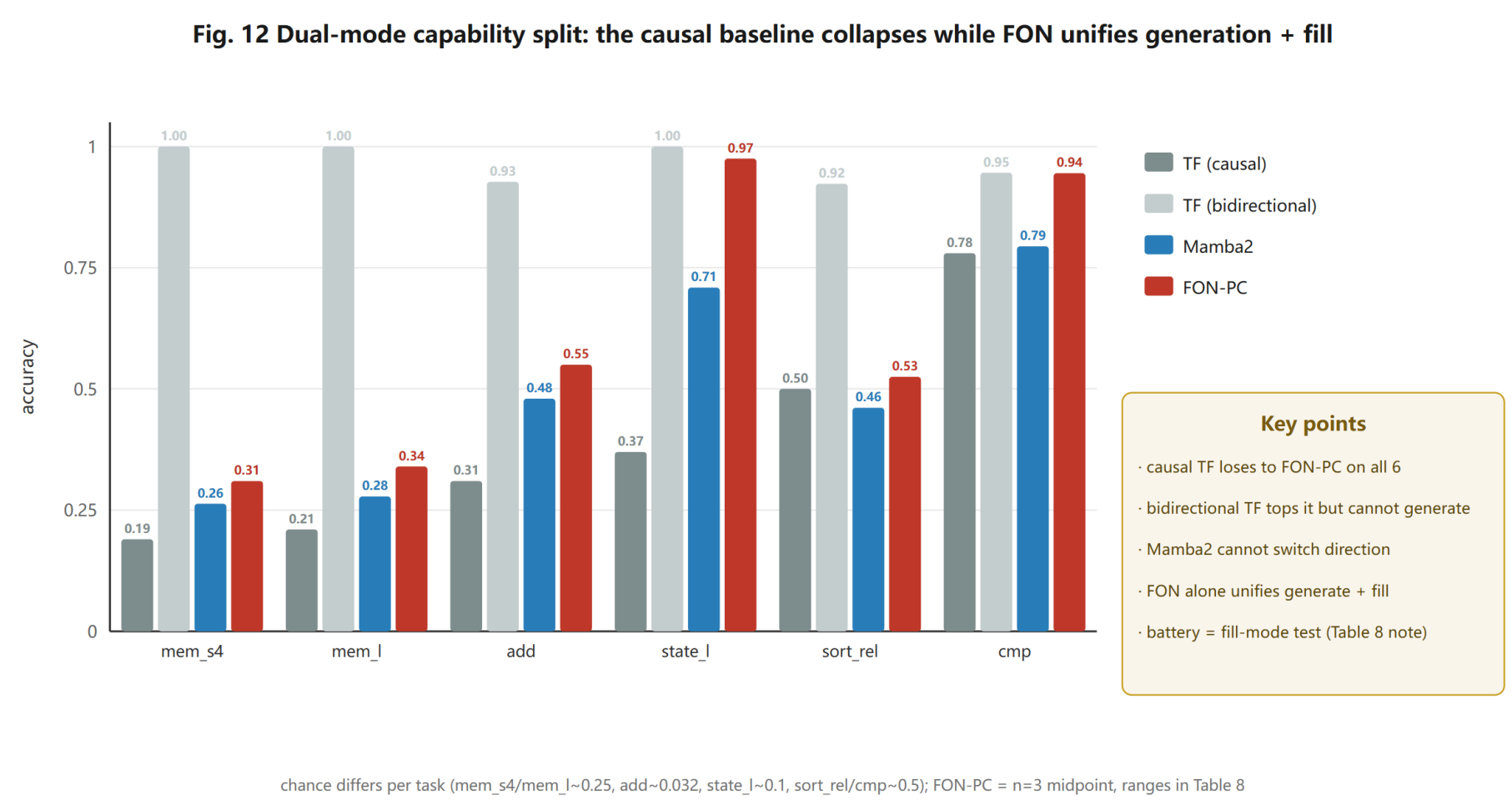

> **Fig. 12.** Dual-mode capability split on the task battery (d256, fill-mode evaluation; FON-PC shows the n=3 midpoint, ranges in Table 8). The causal Transformer—the standard autoregressive baseline—collapses and is outperformed by FON-PC on all six tasks; the bidirectional Transformer tops the battery but cannot generate; Mamba2 cannot switch direction. FON alone unifies causal generation with bidirectional fill.

### 4.6 RQ5: Design degrees of freedom

**Converged degrees of freedom (Fig. 10).** (a) gamma addressing dose 0→0.20 all flat (299.18/299.20/298.87/298.91) [d128·not verified at d256]—module dropped; (b) addressing window W=32 has an inverted-U optimum (299.16/298.87/299.27) [d128]; (c) position kernel × ripple dose 2×2 (d256, T1 √): (posA, 0.02)=131.03 / (posA, 0.005)=131.94 / (pos0, 0.005)=135.30 / (pos0, 0.02)=135.44—position-kernel removal costs +2.6~+3.4% (consistent across both doses, beyond the noise band); dose variation +0.1~+0.7% (within the noise band), no significant PPL effect; (d) coupling-position differences <0.4% (d128), a pseudo-degree of freedom. Additionally: ripple depth K=2 saturates [d128]; the 0.27M ripple stack beats a 17× larger full block [d128]; kernel widths (8,32) are flat [d128].

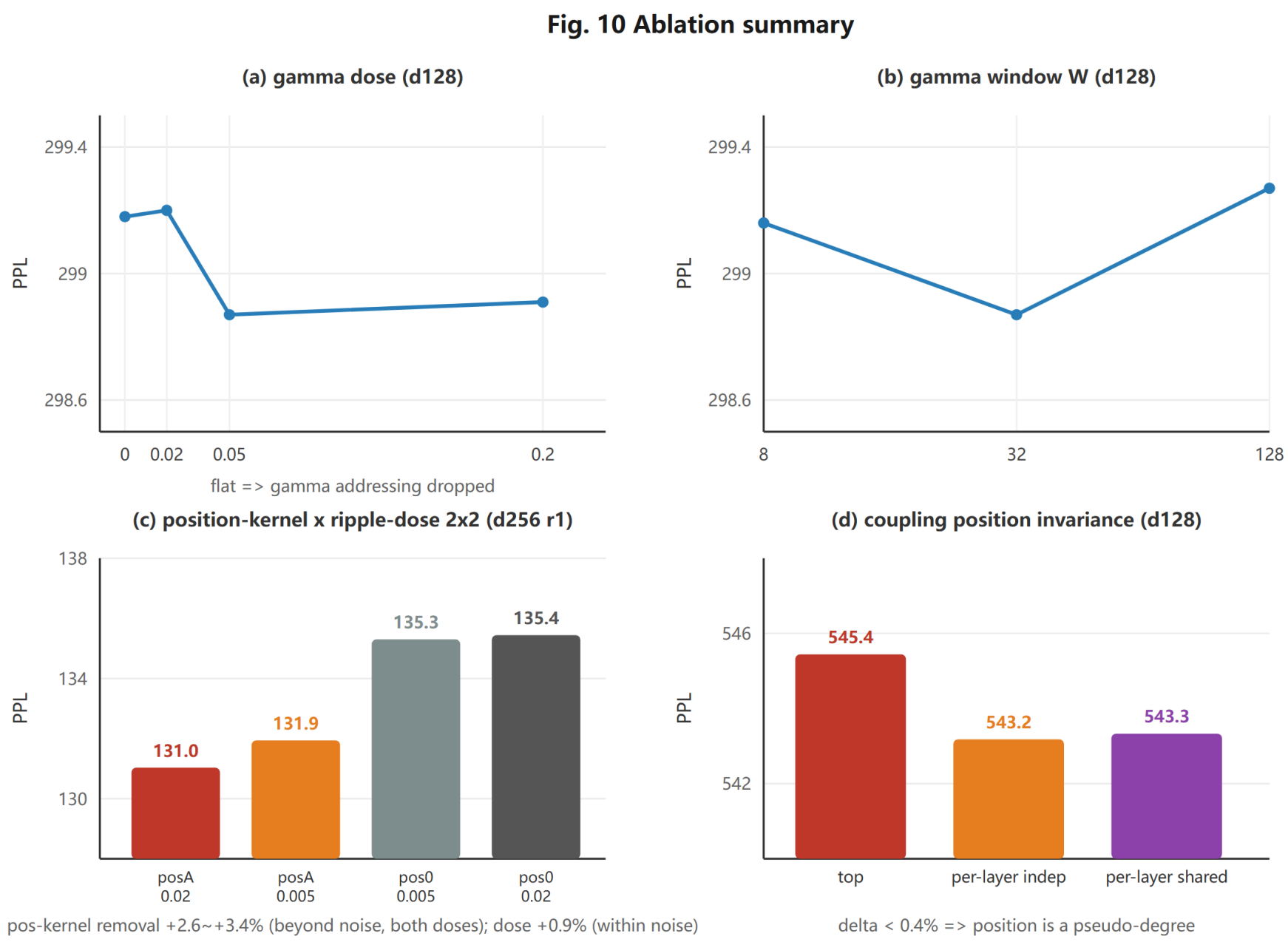

**Fig. 10.** Ablation summary. (a) gamma dose (d128); (b) gamma window (d128); (c) position kernel × dose 2×2 (d256, T1); (d) coupling position (d128).

**Scale reversal (Fig. 11).** Removing the position kernel helps at d128 (−0.27%) but hurts at d256 (+2.6%~+3.4%); another new architectural variant shows −0.54% at d128, +2.56% at d256, +3.13% at d512. **Small-scale conclusions must not be extrapolated**—all [d128]-marked conclusions in this paper are subject to this constraint.

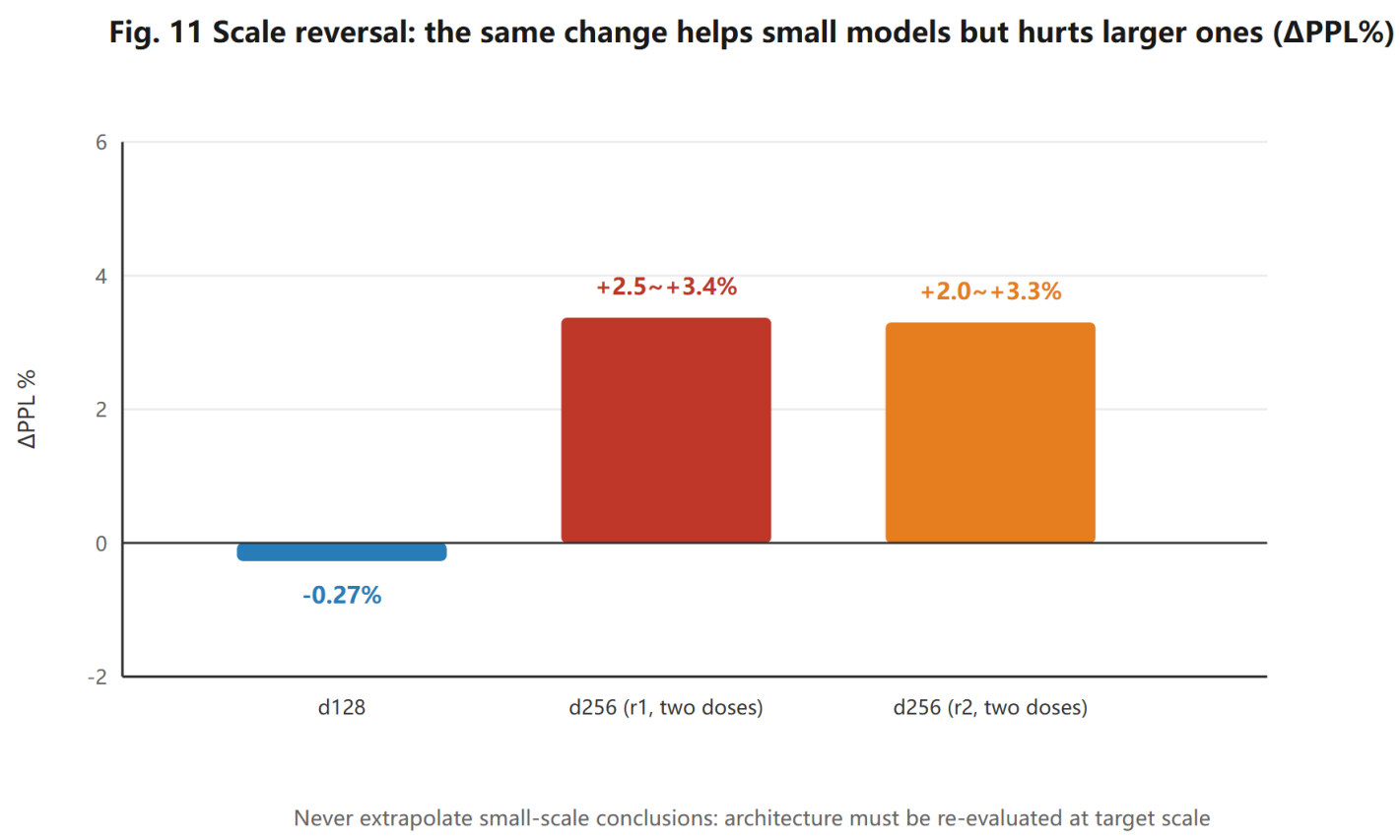

**Fig. 11.** Scale reversal: ΔPPL of position-kernel removal (d128 −0.27% vs. d256 +2.6%~+3.4%).

**The mechanistic explanation of "why FON is better"** is future work: visualizations of the learned delay-tap distribution $\{d_m\}$ and the content-adaptive forgetting $\lambda_m(x)$ after training, and NLL decompositions by position/word frequency/context length, are not covered here.

---

## 5 Discussion

**The boundary between advantage axes and disadvantage axes.** All results in this paper reduce to one load–architecture correspondence law: the four axes on which FON dominates (natural-corpus PPL, engineering-free extrapolation, in-domain infilling, O(1) decode state) correspond to the normal workloads of language modeling. The synthetic-algorithmic battery—once read with the causal-mode control (Table 8)—is **not** a fifth Transformer advantage: it is a fill-mode test that the causal (autoregressive) Transformer fails outright, and only the non-generative bidirectional Transformer tops it (retained deliberately as a ceiling reference, not a generative competitor). The one axis on which attention genuinely leads FON is out-of-domain code infilling (local exact copying; §4.4 Note (4)). Architecture choice is therefore a function of the workload profile, not a contest of single-point scores.

**The revision organ's existential defense.** The drafting-only organ at equal compute is indistinguishable from FON-PC in perplexity within measurement precision (127.54 vs. 126.75), and long-training gains are attributed to recursion thickness (Table 6). The revision organ has exactly one value, but with hard evidence: **the capability envelope**—in exchange for a perplexity cost indistinguishable within measurement precision, it buys the bidirectional revision channel that causal architectures lack in principle; its value holds for in-domain infilling (+19% over on-demand coupling, +24% over drafting-only) and the 0-vs-1 structural-exclusion evidence, and fails for out-of-domain code infilling (§4.4 Note (4)). Its costs (retrieval impairment, +1.1% parameters, +31% mixed-stack FLOPs over Transformer (1.31×)) are fully quantified. Under the operational definition of §1.4, the revision organ satisfies all three criteria of a functional organ (independent parameter block; mask-switchable; selective loss of infilling when switched off).

**List of conclusions that hold only at small scale** (consistent with the scale-reversal warning of Fig. 11): the lambda probe, gamma dose/window, coupling-position invariance, the gating-dose plateau, K=2 saturation, "small and accurate"—all are d128-tier, not verified at d256.

**Limitations.** (1) Scale ceiling: two tiers, 30M (d256) and 70M (d512), sequences ≤2048, single corpus; **the 300M–500M tier is not verified** (this paper does not cover that tier and explicitly lists it as future work)—"the advantage narrows but persists" is currently a two-point observation, and linear extrapolation to larger scales is not supported by our data; (2) the undertrained regime (2.7 tokens/param); (3) a cure for the synthetic retrieval impairment (exclusionary evidence given; positive causal evidence is future work); (4) wall clock needs a custom kernel; (5) LAMBADA is uninformative at the 30M scale (acc ≈ 0.1%) [62]; (6) out-of-domain infilling is weaker than a same-recipe causal Transformer (§4.4 Note (4)).

**Delimitation from the nearest mechanisms.** The Bicameral Model [24] (independent dual models + task supervision) and ResonatorLM [25] (single organ) are delimited in §2.4.

**Future directions.** Scaling; the dose Pareto front and cure of the retrieval impairment; an inference-time review loop for the revision organ; mechanistic explanations (delay-tap visualization).

### Concluding remarks

At the 30M–70M undertrained tier we establish one architectural fact: a single weight stack explicitly differentiated into drafting and revising organs outperforms Transformer and Mamba2 under iso-parameter and iso-token budgets, and outperforms every depth-aligned control, while keeping an O(1) decode state and extrapolating without positional engineering. The revision organ buys capability, not score. The two unknowns that would most change the picture are the 300M+ scale point and a cure for the retrieval impairment; we have stated in advance how each outcome would be adjudicated.

---

## 7 Methods

**Model.** Drafting organ: L=4 layers, d=256/512, learnable delay taps n=16 + content-adaptive recursion (SelCADO, λ zero-initialized), trunk recurs r∈{1,2} rounds, vector readout, position kernel retained. Revision organ: K=2 ripple stack (kernel widths 8/32, 0.266M at d256) + gamma content addressing (0.066M, hole-activated, k/v from the token-embedding layer), gate initialized at 0.02. Embedding and output head shared (V=50304).

**Layer equations.** Drafting-organ mixing operator: $S[p] = \Sigma_m \lambda_m(x[p])\cdot S[p-d_m] + B(x[p])\cdot x[p]$, $\lambda_m(x)=\exp(-\text{softplus}(\rho_m+W_\lambda\ x[p]))$, $y[p]=C(x[p])\cdot S[p]$; revision organ: ripple(h)[p] = Σ (s∈{8,32}) $w_s$·Conv_s(LN(h))[p] (causal kernels at non-hole positions, bidirectional kernels at holes); addressing: $\text{inject}_p = \gamma\cdot W_o(\text{ctx}_p)\cdot 1[\text{hole}]$; fusion: h ← h + gate ⊙ mix(ripple, γ·addr).

**Data and training.** WikiText-103 [8]; formal convention p_synth=0.35 (chosen and frozen early in project development); AdamW (β=(0.9, 0.95), wd 0.1, warmup 300, cosine decay, gradient clipping 1.0), lr 3e-4, batch 16, 20k steps, fp32; hole finetuning lr 1e-4, 4k steps, p_fill=0.5, self-holing stride-8 p=0.15. Hardware: 2× NVIDIA RTX A6000 48GB; all experiments (including exploration and ablations) total several hundred GPU-hours, with per-arm durations available in the open logs.

**Cross-corpus transfer and generation quality (g08 chain, d256).** Cross-corpus PPL (evaluation sequence length 256; domain-shifted vocabulary mismatch inflates all magnitudes, so only horizontal comparison is meaningful): c4—FON-PC-T1 2773 vs. Transformer 3467 vs. Mamba2 3251; code—10005 vs. 15861 vs. 19513; agnews—7948 vs. 10642 vs. 11871 (all models degrade out of domain, but FON is consistently best). Generation quality: distinct-2 = 0.217 (FON-PC-T1) / 0.236 (FON-PC-T2) vs. Transformer 0.179 vs. Mamba2 0.145; repeat-4gram comparable across models (0.56–0.61) [65].

**Evaluation protocols.** PPL (causal, validation set); fill (full-vocabulary argmax, chance=1/50304≈0.002%, key baseline being the same-architecture no-hole model; hole generation: task-given holes / stride-8 self-holing p=0.15); a 25-task battery (synthetic algorithmic skills [66], chance defined per task; **evaluated in fill mode with the answer at a MASK position—for the Transformer this is full bidirectional attention, and a causal-mask mode control on the same checkpoint is reported in Table 8**); length extrapolation (direct evaluation 256→2048 [64]); cross-corpus transfer (agnews/arxiv/c4/code); code FIM (baseline-subtracted); generation quality (distinct-n / repeat-4gram [65]); the evaluation framework follows [63].

**Statistics.** Between-group differences are judged by |Δ| against the noise band (d256 ±2.34%, population-standard-deviation convention); single-seed results support conclusions only when beyond the noise band; main results use n=3 (seeds 0/1/2, per-seed raw values in the appendix); for "indistinguishable" conclusions (e.g., the revision organ's PPL contribution ≈ 0) the equivalence boundary is declared in advance as the noise band.

**Availability.** Data-availability, code-availability, author-contribution and competing-interest declarations are given at the end of the manuscript.

---

## Author contributions

Helong Hu: conceptualization, architecture design, method implementation, experiments, data analysis, manuscript writing and revision. Hongdan Pan: literature survey, framework argumentation, manuscript structuring and polishing. Shuiqing Hu: data curation, result verification, and proofreading. All authors discussed and endorse the conclusions.

## Competing interests

The authors declare no competing interests.

## Data and code availability

The evaluation results supporting every table and figure in this paper (per-seed raw values of the main results, ablations, extrapolation, infilling, the task battery, and wall-clock measurements) are released as structured JSON with the paper; training hyperparameters and evaluation protocols are in §7. The experimental code (pretraining, hole finetuning, all evaluation and figure scripts) and reproduction configurations will be open-sourced. Remaining exploratory intermediate data for ongoing follow-up studies are outside this release; full training logs can be provided to the editorial office on request during review. Model weights are available upon reproduction request.

## References

1. [1] Vaswani, A. et al. Attention is all you need. In *Advances in Neural Information Processing Systems* **30** (2017).
2. [2] Bahdanau, D., Cho, K. & Bengio, Y. Neural machine translation by jointly learning to align and translate. In *Proc. 3rd International Conference on Learning Representations* (2015).
3. [3] Hochreiter, S. & Schmidhuber, J. Long short-term memory. *Neural Comput.* **9**, 1735–1780 (1997).
4. [4] Kaplan, J. et al. Scaling laws for neural language models. arXiv:2001.08361 (2020).
5. [5] Hoffmann, J. et al. Training compute-optimal large language models. In *Advances in Neural Information Processing Systems* **35** (2022).
6. [6] Brown, T. B. et al. Language models are few-shot learners. In *Advances in Neural Information Processing Systems* **33** (2020).
7. [7] Radford, A. et al. Language models are unsupervised multitask learners. *OpenAI Technical Report* (2019).
8. [8] Merity, S., Xiong, C., Bradbury, J. & Socher, R. Pointer sentinel mixture models. In *Proc. 5th International Conference on Learning Representations* (2017).
9. [9] Gu, A., Goel, K. & Ré, C. Efficiently modeling long sequences with structured state spaces. In *Proc. 10th International Conference on Learning Representations* (2022).
10. [10] Gu, A. & Dao, T. Mamba: linear-time sequence modeling with selective state spaces. arXiv:2312.00752 (2023).
11. [11] Dao, T. & Gu, A. Transformers are SSMs: generalized models and efficient algorithms through structured state space duality. In *Proc. 41st International Conference on Machine Learning* (2024).
12. [12] Peng, B. et al. RWKV: reinventing RNNs for the transformer era. In *Findings of the Association for Computational Linguistics: EMNLP 2023* (2023).
13. [13] Sun, Y. et al. Retentive network: a successor to transformer for large language models. arXiv:2307.08621 (2023).
14. [14] Beck, M. et al. xLSTM: extended long short-term memory. In *Advances in Neural Information Processing Systems* **37** (2024).
15. [15] Poli, M. et al. Hyena hierarchy: towards larger convolutional language models. In *Proc. 40th International Conference on Machine Learning* (2023).

16. [16] Katharopoulos, A., Vyas, A., Pappas, N. & Fleuret, F. Transformers are RNNs: fast autoregressive transformers with linear attention. In *Proc. 37th International Conference on Machine Learning* (2020).
17. [17] Choromanski, K. et al. Rethinking attention with performers. In *Proc. 9th International Conference on Learning Representations* (2021).
18. [18] Yang, S., Wang, B., Shen, Y., Panda, R. & Kim, Y. Gated linear attention transformers with hardware-efficient training. In *Proc. 41st International Conference on Machine Learning* (2024).
19. [19] Yang, S., Kautz, J. & Hatamizadeh, A. Gated delta networks: improving Mamba2 with delta rule. arXiv:2412.06464 (2024).
20. [20] Qin, Z. et al. Hierarchically gated recurrent neural network for sequence modeling. In *Advances in Neural Information Processing Systems* **36** (2023).
21. [21] Arora, S. et al. Simple linear attention language models balance the recall-throughput tradeoff. In *Proc. 41st International Conference on Machine Learning* (2024).
22. [22] De, S. et al. Griffin: mixing gated linear recurrences with local attention for efficient language models. arXiv:2402.19427 (2024).
23. [23] Lieber, O. et al. Jamba: a hybrid transformer-Mamba language model. arXiv:2403.19887 (2024).
24. [24] Flamant, C., Ghai, U. & Shimizu, K. The bicameral model: bidirectional hidden-state coupling between parallel language models. arXiv:2605.11167 (2026).
25. [25] Chaudhury, A. ResonatorLM: causal resonant field mixing for efficient long-context language modeling. arXiv:2607.05583 (2026).
26. [26] Dong, Y. et al. FANformer: improving large language models through effective periodicity modeling. In *Proc. 42nd International Conference on Machine Learning* (2025).
27. [27] Raposo, D. et al. Mixture-of-depths: dynamically allocating compute in transformer-based language models. arXiv:2404.02258 (2024).
28. [28] Sun, Y. et al. Learning to (learn at test time): RNNs with expressive hidden states. arXiv:2407.04620 (2024).
29. [29] Ye, T. et al. Differential transformer. arXiv:2410.05258 (2025).
30. [30] Zhang, Z. Comprehension without competence: architectural limits of LLMs in symbolic computation and reasoning. arXiv:2507.10624 (2025).
31. [31] Su, J. et al. RoFormer: enhanced transformer with rotary position embedding. arXiv:2104.09864 (2021).
32. [32] Press, O., Smith, N. A. & Lewis, M. Train short, test long: attention with linear biases enables input length extrapolation. In *Proc. 10th International Conference on Learning Representations* (2022).
33. [33] Chen, S., Wong, S., Chen, L. & Tian, Y. Extending context window of large language models via positional interpolation. arXiv:2306.15595 (2023).
34. [34] Peng, B. et al. YaRN: efficient context window extension of large language models. In *Proc. 12th International Conference on Learning Representations* (2024).
35. [35] Kazemnejad, A., Padhi, I., Natesan Ramamurthy, K., Das, P. & Reddy, S. The impact of positional encoding on length generalization in transformers. In *Advances in Neural Information Processing Systems* **36** (2023).
36. [36] Dai, Z. et al. Transformer-XL: attentive language models beyond a fixed-length context. In *Proc. 57th Annual Meeting of the Association for Computational Linguistics* (2019).
37. [37] Devlin, J., Chang, M.-W., Lee, K. & Toutanova, K. BERT: pre-training of deep bidirectional transformers for language understanding. In *Proc. NAACL-HLT 2019* (2019).
38. [38] Du, Z. et al. GLM: general language model pretraining with autoregressive blank infilling. In *Proc. 60th Annual Meeting of the Association for Computational Linguistics* (2022).
39. [39] Fried, D. et al. InCoder: a generative model for code infilling and synthesis. arXiv:2204.05999 (2022).
40. [40] Donahue, C., Lee, M. & Liang, P. Enabling language models to fill in the blanks. In *Proc. 58th Annual Meeting of the Association for Computational Linguistics* (2020).
41. [41] Bavarian, M. et al. Efficient training of language models to fill in the middle. arXiv:2207.14255 (2022).
42. [42] Fedus, W., Goodfellow, I. & Dai, A. M. MaskGAN: better text generation via filling in the blank. In *Proc. 6th International Conference on Learning Representations* (2018).
43. [43] Nie, S. et al. Large language diffusion models. arXiv:2502.09992 (2025).
44. [44] Graves, A., Wayne, G. & Danihelka, I. Neural Turing machines. arXiv:1410.5401 (2014).
45. [45] Graves, A. et al. Hybrid computing using a neural network with dynamic external memory. *Nature* **538**, 471–476 (2016).

46. [46] Weston, J., Chopra, S. & Bordes, A. Memory networks. In *Proc. 3rd International Conference on Learning Representations* (2015).
47. [47] Khandelwal, U., Levy, O., Jurafsky, D., Zettlemoyer, L. & Lewis, M. Generalization through memorization: nearest neighbor language models. In *Proc. 8th International Conference on Learning Representations* (2020).
48. [48] Lewis, P. et al. Retrieval-augmented generation for knowledge-intensive NLP tasks. In *Advances in Neural Information Processing Systems* **33** (2020).
49. [49] Borgeaud, S. et al. Improving language models by retrieving from trillions of tokens. In *Proc. 39th International Conference on Machine Learning* (2022).
50. [50] Shinn, N. et al. Reflexion: language agents with verbal reinforcement learning. In *Advances in Neural Information Processing Systems* **36** (2023).
51. [51] Madaan, A. et al. Self-refine: iterative refinement with self-feedback. In *Advances in Neural Information Processing Systems* **36** (2023).
52. [52] Welleck, S. et al. Generating sequences by learning to self-correct. In *Proc. 11th International Conference on Learning Representations* (2023).
53. [53] Khairi, A., D'souza, D., Fadaee, M. & Kreutzer, J. Making, not taking, the best of N. arXiv:2510.00931 (2025).
54. [54] Giannou, A. et al. Looped transformers as programmable computers. In *Proc. 40th International Conference on Machine Learning* (2023).
55. [55] Jaynes, J. *The Origin of Consciousness in the Breakdown of the Bicameral Mind* (Houghton Mifflin, 1976).
56. [56] Lamme, V. A. F. & Roelfsema, P. R. The distinct modes of vision offered by feedforward and recurrent processing. *Trends Neurosci.* **23**, 571–579 (2000).
57. [57] Lamme, V. A. F. Towards a true neural stance on consciousness. *Trends Cogn. Sci.* **10**, 494–501 (2006).
58. [58] McClelland, J. L., McNaughton, B. L. & O'Reilly, R. C. Why there are complementary learning systems in the hippocampus and neocortex. *Psychol. Rev.* **102**, 419–457 (1995).
59. [59] Kumaran, D., Hassabis, D. & McClelland, J. L. What learning systems do intelligent agents need? *Trends Cogn. Sci.* **20**, 512–534 (2016).
60. [60] Grossberg, S. How does a brain build a cognitive code? *Psychol. Rev.* **87**, 1–51 (1980).
61. [61] Dehaene, S. & Changeux, J.-P. Experimental and theoretical approaches to conscious processing. *Neuron* **70**, 200–227 (2011).
62. [62] Paperno, D. et al. The LAMBADA dataset: word prediction requiring a broad discourse context. In *Proc. 54th Annual Meeting of the Association for Computational Linguistics* (2016).
63. [63] Gao, L. et al. A framework for few-shot language model evaluation. https://github.com/EleutherAI/lm-evaluation-harness (2023).
64. [64] Hsieh, C.-P. et al. RULER: what's the real context size of your long-context language models? arXiv:2404.06654 (2024).
65. [65] Holtzman, A., Buys, J., Du, L., Forbes, M. & Choi, Y. The curious case of neural text degeneration. In *Proc. 8th International Conference on Learning Representations* (2020).
66. [66] Srivastava, A. et al. Beyond the imitation game: quantifying and extrapolating the capabilities of language models. *Trans. Mach. Learn. Res.* (2023).
67. [67] Elhage, N. et al. In-context learning and induction heads. *Transformer Circuits Thread* (2022).
68. [68] Tiezzi, M. et al. Back to recurrent processing at the crossroad of transformers and state-space models. *Nat. Mach. Intell.* **7**, 678–688 (2025).
69. [69] Du, C. et al. Human-like object concept representations emerge naturally in multimodal large language models. *Nat. Mach. Intell.* **7** (2025).